\documentclass{article}
\usepackage{ijcai25}

\usepackage{times}
\usepackage{soul}
\usepackage{url}
\usepackage[hidelinks]{hyperref}
\usepackage[utf8]{inputenc}
\usepackage[small]{caption}
\usepackage{graphicx}
\usepackage{amsmath}
\usepackage{amsthm}
\usepackage{booktabs}
\usepackage{algorithmic}
\usepackage[switch]{lineno}

\usepackage{amsmath,amsfonts,bm}

\def\eqref#1{equation~\ref{#1}}

\def\1{\bm{1}}

\DeclareMathAlphabet{\mathsfit}{\encodingdefault}{\sfdefault}{m}{sl}
\SetMathAlphabet{\mathsfit}{bold}{\encodingdefault}{\sfdefault}{bx}{n}

\usepackage{url}
\usepackage{pgfplots}
\usepackage{pgfplotstable}
\usetikzlibrary{patterns} 
\usepackage{amsmath, amssymb}
\usepackage{tcolorbox}
\usepackage{graphicx}
\usepackage{enumitem}
\usepackage{wrapfig}
\usepackage{subcaption} 

\usepackage{xcolor}

\usepackage{amsmath}
\usepackage{graphicx}
\usepackage{amsmath}
\usepackage{amssymb}

\usepackage{amsfonts}
\usepackage{multirow}
\usepackage{array}

\usepackage{listings}

\RequirePackage{comment}

\pgfplotsset{compat=1.17}
\definecolor{darkgreen}{rgb}{0, 0.6, 0}

\usepackage{fvextra}

\DefineVerbatimEnvironment{Verbatim}{Verbatim}{breaklines=true, breaksymbol={}}

\title{Instantiation-based Formalization of Logical Reasoning Tasks using \\ Language Models and Logical Solvers}

\author{
Mohammad Raza
\And
Natasa Milic-Frayling
\affiliations
Qatar Computing Research Institute
}

\begin{document}

\maketitle
\begin{abstract}
 Robustness of reasoning remains a significant challenge for large language models, and addressing it is essential for the practical applicability of AI-driven reasoning systems. We introduce Semantic Self-Verification (SSV), a novel approach that addresses the key challenge in combining language models with the rigor of logical solvers: to accurately formulate the reasoning problem from natural language to the formal language of the solver. SSV uses a consistency-based approach to produce strong abstract formalizations of problems using concrete instantiations that are generated by the model and verified by the solver. In addition to significantly advancing the overall reasoning accuracy over the state-of-the-art, a key novelty that this approach presents is a feature of verification that has near-perfect precision over a significant coverage of cases, as we demonstrate on open reasoning benchmarks. We propose such \emph{near-certain reasoning} as a new approach to reduce the need for manual verification in many cases, taking us closer to more dependable and autonomous AI reasoning systems.
\end{abstract}

\section{Introduction}
\label{sec:intro}

Logical reasoning remains a persistent challenge for large language models (LLMs). Although these models demonstrate reasoning capabilities across various domains, their reasoning often lacks robustness and becomes increasingly error-prone as task complexity increases. Many recent approaches have made notable advancements in this active area of research. Chain-of-thought (CoT) prompting has demonstrated how the quality of reasoning can be improved by prompting the model to explicitly generate the steps of reasoning in natural language before arriving at the final answer \cite{cot}. Variants of CoT and other related prompting and fine-tuning approaches have shown further improvements \cite{least-to-most,self-consistency,thought-propagation,self-verification-emnlp23,selection-inference}. To address the logical inconsistencies that can arise in such natural language approaches, another interesting direction is to  incorporate LLMs with logical solvers or automated reasoning tools \cite{logic-lm,sat-lm}. Rather than directly attempting reasoning with the LLM, these approaches use the LLM to infer a formal representation of the problem as a program that can be executed by the solver, as such automated reasoning tools guarantee logically sound inference by construction. 

While these approaches have demonstrated relative improvements in accuracy, we are still far from achieving robustness and reliability of reasoning. For instance, Figure \ref{fig:mot-example_a} shows an example reasoning problem from the Law School Admissions Test on analytical reasoning \cite{ar-lsat}. On tasks of such complexity, the best reported accuracy, achieved by a solver-augmented system, is only 43\% \cite{logic-lm}. Such lack of reliability especially hinders the practical usability of existing approaches: the burden of verifying correctness is \emph{always} on the user, which can be especially difficult and error-prone for complex reasoning tasks. Therefore, having a reliable signal of correctness with high confidence can be hugely beneficial to help reduce the overall manual effort and cost of verification. 

In this work, we propose a new approach to correctly formalizing reasoning problems called \emph{Semantic Self-Verification} (SSV), which offers two key benefits: (1) it improves the overall accuracy of reasoning significantly over SoTA, and (2) it provides a novel feature of verification that has \emph{near-perfect} precision. In our problem formulation, in addition to producing an answer to a given question, the system also indicates if it was able to \emph{verify} the correctness of the answer: {\small $\textsf{Question} \rightarrow (\textsf{Answer}, \textsf{isVerified})$}. This problem formulation is similar to confidence estimation in machine learning, except that in our case the $\small\textsf{isVerified}$ indicator is a boolean rather than continuous value: if true, it indicates a ``near certain" confidence in the correctness of the answer. Such high-confidence verification  can reduce the need for manual checking in many cases.

    \begin{figure}[t]
        \centering
        \begin{tcolorbox}[colframe=black, colback=gray!10, boxrule=0.2mm]

        \textsf{\tiny In a repair facility, there are exactly six technicians: Stacy, Urma, Wim, Xena, Yolanda, and Zane. Each technician repairs machines of at least one of the following three types—radios, televisions, and VCRs—and no other types. The following conditions apply: Xena and exactly three other technicians repair radios. Yolanda repairs both televisions and VCRs. Stacy does not repair any type of machine that Yolanda repairs. Zane repairs more types of machines than Yolanda repairs. Wim does not repair any type of machine that Stacy repairs. Urma repairs exactly two types of machines. Which one of the following pairs of technicians could repair all and only the same types of machines as each other?}
\\[0.7em]
        \textsf{\tiny {\bf\textsf{(A)}} Stacy \& Urma \hspace{1em} \\{\bf\textsf{(B)}} Urma \& Yolanda \hspace{1em} \\ {\bf\textsf{(C)}} Urma \& Xena \hspace{1em} \\{\bf\textsf{(D)}} Wim \& Xena \hspace{1em} \\ {\bf\textsf{(E)}} Xena \& Yolanda}
        \end{tcolorbox}
        \caption{Sample problem from the Law School Admissions Test}
        \label{fig:mot-example_a}
    \end{figure}
    
    \begin{figure}[t]
        \centering
        
        \includegraphics[width=0.48\textwidth]{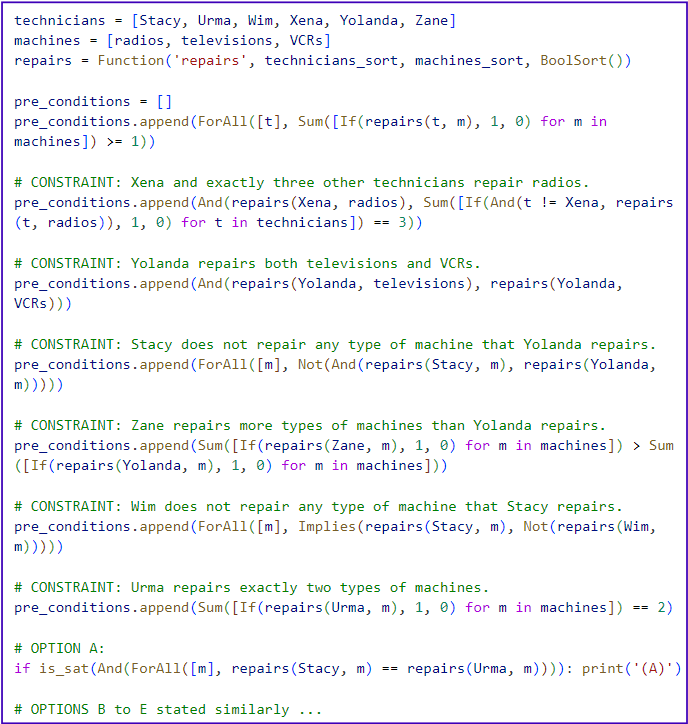} 
        
        \caption{Sample problem formalization as Z3 code}
        \label{fig:mot-example_b}
    \end{figure}

At its core, our approach addresses the key challenge in combining LLMs with the robust reasoning of logical solvers: the formulation of a problem from informal natural language (NL) to the formal representation that is a program executable by the solver. For example, Figure \ref{fig:mot-example_b} shows the formal representation of the NL problem from Figure \ref{fig:mot-example_a}. In this case the formalization is expressed as code in the language of the Z3 SMT solver \cite{z3}, which is a state-of-the-art industrial strength theorem prover that can produce the correct answer when given these correctly-expressed formal constraints. The crucial task, therefore, is for the LLM to correctly translate the NL problem description to such a formal representation, and this is where LLMs can make significant errors, as shown by the limits of prior work \cite{logic-lm,sat-lm}. 

Our approach of verifying that a formal representation is true to the original problem is inspired by how humans often create formalizations of problems expressed in natural language. For instance, when school students solve math word problems, they must first create the right algebraic equation that represents the problem, before they can solve it to get the answer. To ensure that their translation to an abstract equation represents the problem correctly, they are encouraged to consider various example instances of the problem and to check that the abstract equation consistently satisfies those instances so that it all ``makes sense". In the same way, in the SSV approach, rather than just doing a single abstract translation from NL to a formal representation, we also use the LLM to additionally generate various \emph{concrete instantiations}, or examples, of the general constraint, which are used as test cases to check the correctness of the abstract formalization. Using the logical solver, we verify that each of these instantiations is consistently satisfied by the formal representation. If all of these distinct semantic relationships  consistently hold, then verification passes.

\begin{figure}
    \centering
    
    \begin{subfigure}[b]{0.42\textwidth}
        \centering
        
        \includegraphics[width=\textwidth]{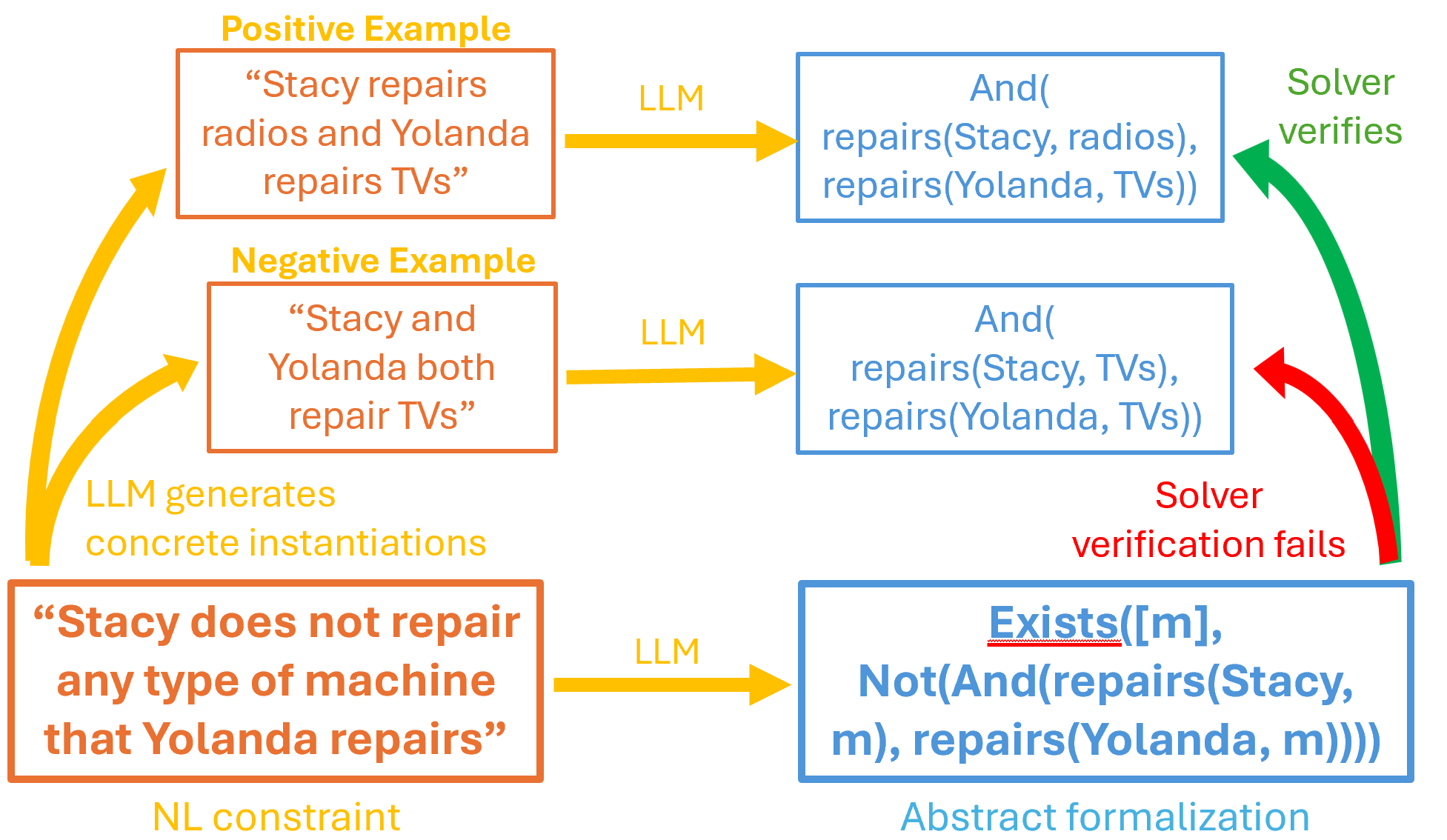} 
        
        \caption{Incorrect formalization (uses $\mathsf{Exists}$  quantifier)}
        \label{fig:ssv-example_a}
    \end{subfigure}
    \hfill
    \begin{subfigure}[b]{0.42\textwidth}
        \centering
        
        \includegraphics[width=\textwidth]{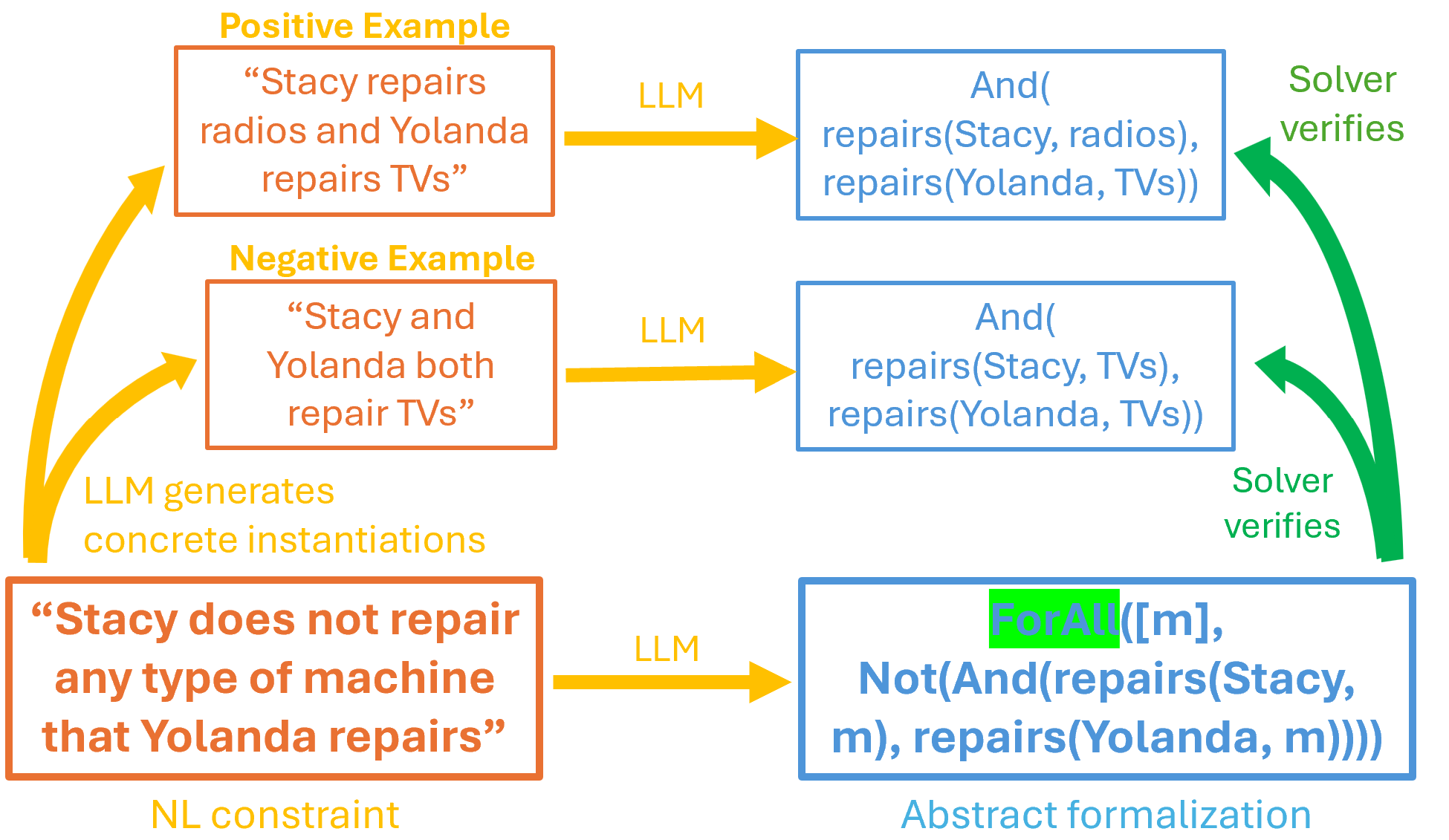} 
        
        \caption{Correct formalization (uses $\mathsf{ForAll}$ quantifier)}
        \label{fig:ssv-example_b}
    \end{subfigure}

    \caption{Semantic self-verification of a general constraint:  the negative example fails for the wrong formalization (a), while both instantiations are verified for the correct formalization (b)}
    \label{fig:ssv-example}
\end{figure}

Figure \ref{fig:ssv-example} illustrates how the SSV approach works for the third constraint from the problem in Figure \ref{fig:mot-example_b}, which requires that Stacy and Yolanda cannot repair the same type of machine. A direct translation using the LLM may produce an incorrect abstract formalization of this constraint as shown in Figure \ref{fig:ssv-example_a}, where the condition  is asserted only \emph{for some} machine rather than \emph{for all} machines because the $\mathsf{Exists}$ quantifier is incorrectly used. However, in the SSV approach, we use the LLM to additionally infer simple concrete instantiations, or examples, of the general constraint. For instance, a concrete positive example is that Stacy  repairs radios and Yolanda repairs TVs. A  negative example is that Stacy and Yolanda both repair TVs. After  inferring these examples in NL, we also use the LLM to translate them to formal expressions in the language of the solver. We then use the solver to check that each of these expressions is consistent with the abstract formalization. In Figure \ref{fig:ssv-example_a} we see that the negative instantiation fails verification because the abstract formalization does not assert the condition for all machine types, so it still allows  Stacy and Yolanda to both repair TVs. However, with the correct formalization in Figure \ref{fig:ssv-example_b} that uses the $\mathsf{ForAll}$ quantifier,  both instantiations pass the solver verification, since the abstract formalization correctly disallows that \emph{any} machine can be repaired by both technicians.

We note that any notion of verification from natural to formal language cannot provide formal correctness guarantees, since natural language itself is inherently informal and often ambiguous. However, as we demonstrate empirically, a passing verification in our case indicates a \emph{near certain} confidence in the answer correctness since multiple independent semantic relationships are consistently satisfied. In this respect, our approach is akin to a consensus-based ensemble as it is based on agreement between multiple independent predictors \cite{ensemble-learning}. However, rather than all predictors addressing \emph{the same} task, we have a \emph{semantic ensemble} of predictors that are addressing different but semantically related tasks and the logical solver verifies the formal consistency between these. We also note that unlike standard proposer-verifier approaches, in our case there is no verifier that can check \emph{correctness} of a proposed formalization: our verification is thus based on formal \emph{consistency} between abstract and concrete inferences. 

Furthermore, having such a high precision verification mechanism also allows us to improve the formalization itself, in two different respects. Firstly, any failing instantiation can be used as concrete guidance to refine the formalization further, as it can hint at potential errors. This is similar to error-based refinement in code generation techniques \cite{error-based-refinement-1}, except that here we are guided by  \emph{semantic} errors inferred from the instantiations rather than just \emph{syntactic} execution errors in the code. Secondly, with our verification mechanism we can also explore the search space more extensively: using temperature sampling to create multiple candidate formalizations and selecting ones that pass verification.

\begin{figure} 
  \centering
  \includegraphics[width=0.48\textwidth]{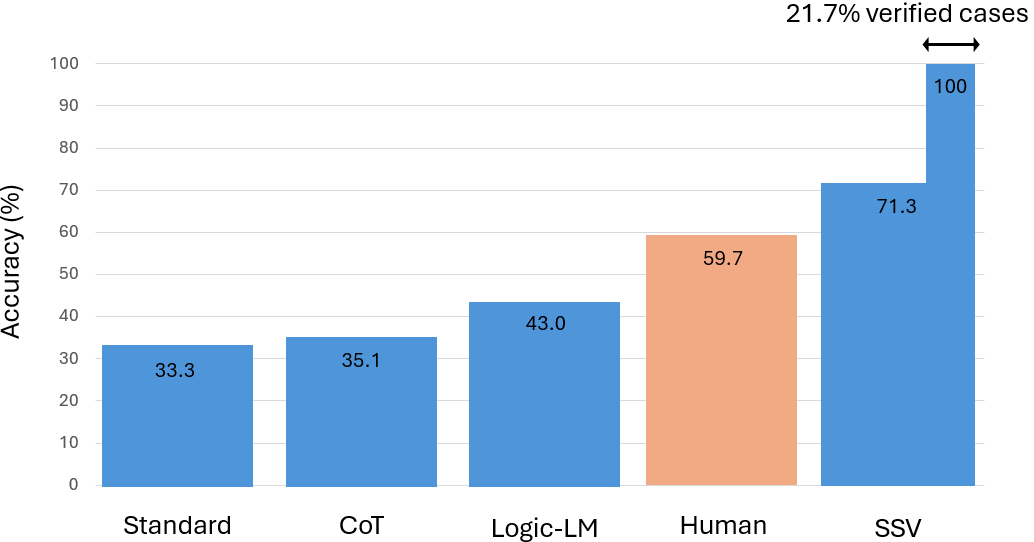} 
  \caption{Towards near-perfect reasoning: SSV achieves new SoTA accuracy and 100\% verification precision on the AR-LSAT law school tests dataset (\emph{all systems using GPT-4 as base LLM}).}
  \label{fig:comparison-barchart}
\end{figure}

Our evaluation demonstrates how the SSV approach achieves a significant increase in overall accuracy, as well as a near-perfect precision (or selective accuracy) on the verified cases. Figure \ref{fig:comparison-barchart} highlights the results for the most challenging AR-LSAT law school tests dataset. Though better than direct LLM inference and CoT, the accuracy of the best performing existing system (the solver-augmented Logic-LM approach by \cite{logic-lm}) is at 43\%, while SSV achieves a significantly higher accuracy of 71.3\%, which also surpasses the average human performance. Moreover, the precision of the 21.7\% of cases that it is able to verify is 100\%. This means that a 21.7\% reduction in manual verification effort can potentially be made on tasks of such high complexity. In our full evaluation we also show higher accuracy and coverage of verified cases on other standard reasoning datasets.

In summary, we make the following contributions in this work: (1) We propose the problem formulation of returning a boolean high-confidence verification indication in addition to the answer, which can be used to reduce manual cost of verification. (2) We present the novel technique of semantic self-verification, which uses concrete instantiations to verify the correctness of the problem formalization. (3) We show how SSV can also improve the formalization itself through instantiation-guided refinement and exploration of multiple candidate formalizations. (4) We present an extensive evaluation on five open benchmarks that shows a significant increase in overall accuracy over SoTA, as well as near-perfect selective accuracy over a significant coverage of verified cases.\footnote{code \& data available at \url{http://github.com/mohammadraza4/ssv}}

\section{Semantic Self-Verification}
\label{sec:ssvalgorithm}

This section describes the semantic self-verification approach for reasoning problems, which generates programs verified and refined by concrete instantiations. Figure \ref{fig:ssv-alg} presents the main algorithm, illustrating the top-level flow and key components. As formulated, the algorithm takes a question ($\mathtt{Q}$), such as the technicians problem in Figure \ref{fig:mot-example_a}, and outputs an answer along with an indication of verification success. Figure \ref{fig:ssv-alg} also details the algorithm's configuration parameters: the chosen LLM and solver, LLM temperature values, and the maximum repair attempts. We first outline the general algorithm before discussing its key phases in detail.

For each temperature value to be explored, the algorithm first uses the LLM to infer a program $P$ that the solver executes to answer the question $\mathtt{Q}$, such as the program from Figure \ref{fig:mot-example_b}. If an executable program is generated ($P \neq \varnothing$), the verification loop begins (line 4). The solver first executes $P$ to obtain an answer. Then, for verification, we infer concrete instantiations $\mathcal{I}$, which are test cases for the program’s constraints and options, such as the six constraints and five options in Figure \ref{fig:mot-example_b}. The solver attempts to verify that each instantiation is formally satisfiable and returns any failing instantiation $I_{\text{fail}}$. For example, for the third constraint in the technicians program, inferred instantiations (Figure \ref{fig:ssv-example_a}) may yield the failing case: ``Stacy and Yolanda cannot both repair TVs.'' If no failing instantiation is found (as in Figure \ref{fig:ssv-example_b}) and $P$ satisfies general well-formedness properties, the algorithm returns its answer $A$ along with verification success (line 12).

If verification fails, we attempt to repair the program $P$ using the LLM and any failing instantiation, which provides insight into potential constraint implementation errors. For example, the failing instantiation in Figure \ref{fig:ssv-example_a} may guide the LLM to assert the condition for all machine types using the forall quantifier, as shown in Figure \ref{fig:ssv-example_b}. After obtaining the repaired program, we repeat the verification loop. If no answer is verified across all temperatures and repair attempts, we exit the outer loop (line 16). If no executable program was inferred, we fall back to direct inference using the LLM with a chain-of-thought prompt, as in prior work \cite{logic-lm}. Otherwise, we return the best answer with verification failure. We next discuss key algorithm phases in more detail.


{\bf Program generation.} The $\mathtt{GenProgram}$ function in Figure \ref{fig:ssv-alg} uses the LLM to generate a solver-executable program for the given problem. A basic implementation relies on a direct LLM prompt, but we incorporate techniques from the code generation literature to improve quality. First, we use error-based refinement: syntax or execution errors in the generated program are fed back to the LLM for repair, a common approach in LLM-based code generation/reasoning domains \cite{error-based-refinement-1,logic-lm}. Second, if direct code generation fails, we employ a compositional approach \cite{decomposed-prompting,din-sql}, generating the program incrementally for each identified constraint. This improves code quality compared to direct prompting, which often produces syntax errors.

{\bf Semantic verification.} While code generation ensures an executable solver program, it does not address \emph{semantic} correctness—whether the program accurately implements the problem's intended constraints. SSV addresses this by generating and verifying concrete instantiations for each constraint in the generated program. The $\mathtt{GenInstantiations}$ function first parses the program $P$ to extract constraints and their NL descriptions. Our program generation phase structures programs in segments of the form $P_{init} + C_1 + ... + C_N + O_1 + ... + O_M$, where $P_{init}$ contains initial definitions, followed by explicitly segmented constraints and options, each annotated with NL comments (e.g. see ``\#CONSTRAINT:'' and ``\#OPTION:'' segments in Figure \ref{fig:mot-example_b}). This structure allows parsing constraints along with their NL descriptions.

\begin{figure}
    \centering
    

    \begin{algorithmic}[1]
    \REQUIRE $\mathtt{Q}$ \qquad  \emph{// the question}
    \REQUIRE $\mathtt{LLM}$ \qquad  \emph{// the language model}
    \REQUIRE $\mathtt{Solver}$ \qquad  \emph{// the logical solver}
    \REQUIRE $\mathtt{Temperatures}$ \quad  \emph{// LLM temperatures to try}
    \REQUIRE $\mathtt{MaxRepairs}$ \quad \emph{// maximum repair attempts}
    
    \STATE $A_{\text{best}} \gets \varnothing$
    \FOR{\bf{each} $T \in \mathtt{Temperatures}$}
        \STATE $P \gets \texttt{GenProgram}(\mathtt{LLM}, T, \mathtt{Solver}, \mathtt{Q})$
        \WHILE{$P \neq \varnothing$ \textbf{and under} $\mathtt{MaxRepairs}$} 
            \STATE $A \gets \texttt{ExecuteProgram}(\mathtt{Solver}, P)$
            \IF{$A_{\text{best}} = \varnothing$}
                \STATE $A_{\text{best}} \gets A$
            \ENDIF
            \STATE $\mathcal{I} \gets \texttt{GenInstantiations}(\mathtt{LLM}, T, P)$
            \STATE $I_{\text{fail}} \gets \texttt{Verify}(\mathtt{Solver}, \mathcal{I}, P)$
            \IF{$I_{\text{fail}} = \varnothing$ \textbf{and} $\texttt{IsWellFormed}(P)$}
                \RETURN $(A, \texttt{True})$
            \ENDIF
            \IF{$A = \varnothing$}
            \STATE $P \gets \texttt{RepairProgram}(\mathtt{LLM}, T, \mathtt{Q}, P, I_{\text{fail}})$
            \ENDIF
        \ENDWHILE
    \ENDFOR
    \IF{$A_{\text{best}} = \varnothing$}
        \STATE $A_{\text{best}} \gets \texttt{InferLLMAnswer}(\texttt{LLM}, \mathtt{Q})$
    \ENDIF
    \RETURN $(A_{\text{best}}, \texttt{False})$
    \end{algorithmic}
    \caption{The Semantic Self-Verification Algorithm}
    \label{fig:ssv-alg}
\end{figure}

We use the LLM to infer concrete instantiations for each of the constraints, using their NL descriptions. For each constraint $C_i$, our implementation prompts the LLM for one positive and one negative instantiation, and both instantiations are translated into solver expressions (Figure \ref{fig:ssv-example}). Once all instantiations $\mathcal{I}$ are obtained, the $\mathtt{Verify}$ function uses the solver to check if each constraint is consistent with its respective instantiations. For each constraint $C_i$, we  it verifies its positive instantiation $I_p$ by constructing and executing the   expression $P_{init} + C_i + I_p$ and checking that the solver returns SAT. For the negative instantiation $I_n$, it checks that the expression $P_{init} + C_i + I_n$ is UNSAT. If this holds for all constraints, the full program is considered verified. If verification fails, it returns the first failing instantiation $I_{\text{fail}} \in \mathcal{I}$. 

Beyond verifying concrete instantiations, we also check general logical well-formedness properties using the $\mathtt{IsWellFormed}$ function, which ensures (1) the program follows the specified structure, (2) it returns a single answer, and (3) it avoids degenerate expressions—tautologies or vacuous implications that introduce redundancies or oversimplifications in the problem formalization.

\begin{table*}[h]
\centering
\small{
\begin{tabular}{c||c|c|c|c||c|c}
\hline
\rule{0pt}{2.5ex}
\multirow{2}{*}{\textbf{Dataset}} & \multicolumn{4}{c||}{\textbf{General Accuracy}} & \multicolumn{2}{c}{\textbf{SSV Verification}} \\
\cline{2-7} 
 & \rule{0pt}{2.5ex} \textbf{Standard} & \textbf{\quad CoT\quad} & \textbf{Logic-LM} & \textbf{\quad SSV\quad} & \textbf{Coverage} & \textbf{Precision} \\
\hline \rule{0pt}{2.5ex}
AR-LSAT  &   33.3  & 35.1  & 43.0 & \bf{71.3} & 21.7 & 94.0 \textcolor{darkgreen}{(100.0)}  \\
\rule{0pt}{2.5ex} FOLIO  & 69.1   & 70.6  & 78.9 & \bf{80.9} & 25.0 & 98.0 \textcolor{darkgreen}{(100.0)}\\
\rule{0pt}{2.5ex} LogicalDeduction & 71.3  & 75.3   & 87.6 & \bf{89.7} & 43.7 & 100.0 \\
\rule{0pt}{2.5ex} PrOntoQA & 77.4  & 98.8  & 83.2 & \bf{100.0} & 66.0 & 100.0 \\
\rule{0pt}{2.5ex} ProofWriter & 52.7  & 68.1  & 79.7 & \bf{98.0} & 75.2 & 98.7 \textcolor{darkgreen}{(100.0)} \\\hline

\end{tabular}
}

\caption{General accuracy and SSV precision/coverage with GPT-4 base model. \emph{Values in brackets are actual values on corrected datasets.}}
\label{tab:eval-general}
\end{table*}


{\bf Semantic program repair.} If  verification fails and a failing instantiation $I_{\text{fail}}$ is found, the $\mathtt{RepairProgram}$ function attempts to repair the original program $P$, provided no answer has been found. Unlike error-based program repair, this is a \emph{semantic} repair based on an instantiation inferred by the LLM rather than an execution error. In our repair prompt, we supply the initial definitions code, the constraint code with its NL description, and the failing instantiation expression. The LLM is prompted to first do a chain-of-thought analysis to infer whether the error lies in the initial definitions, the constraint code, or the instantiation itself, before inferring the corrected code. The prompts used for code generation/refinement, instantiation generation and semantic repair are available in the appendix.

\section{Evaluation}
\label{sec:evaluation}


We evaluate our SSV technique on open benchmarks for logical reasoning, focusing on two key aspects: (1) improving the general accuracy of reasoning over existing baselines and (2) assessing verification quality in terms of both precision (correctness) and coverage (proportion of verified cases).

\emph{Datasets}. We use five common datasets for logical reasoning. All datasets follow a multiple-choice format, where each task includes a problem statement, a question, and answer options (e.g., Figure \ref{fig:mot-example_a}). {\bf{PrOntoQA}} is a synthetic deductive reasoning dataset for LLM evaluation \cite{pronto-qa}. We use its most challenging subset—fictional character tasks requiring 5 reasoning hops—comprising 500 test examples with 2 answer options (True/False). {\bf{ProofWriter}} is a widely used logical reasoning dataset \cite{proof-writer}. We use its open-world assumption subset with 5-hop reasoning tasks, following \cite{logic-lm}, with 600 test examples and 3 answer options (True/False/Unknown). {\bf{FOLIO}} is an expert-crafted dataset for logical reasoning \cite{folio}, featuring real-world knowledge problems phrased in natural language and requiring complex first-order logic. We evaluate on its full test set of 204 examples, each with 3 answer options (True/False/Unknown). {\bf{LogDeduction}} is a dataset from the BigBench benchmark \cite{log-deduction} involving object sequence ordering based on given conditions. The full test set contains 300 tasks with 3, 5, or 7 answer options. {\bf{AR-LSAT}} consists of analytical reasoning questions from LSAT exams from 1991–2016 \cite{ar-lsat}. This challenging dataset has seen only marginally better-than-random accuracy from existing approaches  \cite{logic-lm,holistic-evaluation-llms}. The test set has 230 questions, each with 5 answer options.


\emph{Baselines}. We compare our technique against three baselines, which represent  approaches of reasoning using the LLM alone, as well as the combination of formal logical solvers with LLMs. Each of these baselines  and our own system is parametric in the LLM used, and in our experiments we investigate all systems with both the GPT-4 model (a current best general LLM for reasoning) as well as the weaker GPT-3.5 model from Open AI. We use the baselines and their results for these models  as reported in \cite{logic-lm}. The baselines are as follows. {\bf Standard} is the direct approach of prompting the LLM, leveraging in-context learning to answer the question. {\bf CoT} (Chain-of-Thought) \cite{cot} follows a step-by-step reasoning process, generating explanations before the final answer. {\bf Logic-LM} is a state-of-the-art method that integrates LLMs with solvers for formal reasoning \cite{logic-lm}, where the LLM is prompted to generate a solver program to solve the task. {\bf SSV} is our semantic self-verification technique (Figure \ref{fig:ssv-alg}). Our implementation uses the Z3 SMT solver \cite{z3} and applies identical prompts for both models, with 1-4 few-shot examples drawn from training datasets (detailed in the Appendices). Our full SSV implementation sets $\mathtt{MaxRepairs} = 2$ and $\mathtt{Temperatures} = [0, 0.3, 0.4, 0.5]$ (covering low to mid-range values), with parameter variations explored in the ablation analysis.

\subsection{Results}


\paragraph{Main results} Table \ref{tab:eval-general} presents the main results, with all systems evaluated using GPT-4 as the underlying LLM. The table reports general accuracy as well as the precision and coverage of SSV verification. General accuracy represents the percentage of correct answers across the dataset. For SSV, precision denotes the percentage of correct answers among those flagged as verified, while coverage indicates the percentage of verified cases relative to the entire dataset. The key observations are as follows:

1. \underline{SSV outperforms all baselines in general accuracy.} Our technique achieves a higher general accuracy over all baseline systems across all datasets. We especially note the drastic increase of 28.3\% over the current best Logic-LM system on the most difficult AR-LSAT dataset. This shows the strong effectiveness of our technique in producing robust problem formalizations in contrast to just a direct LLM translation from the natural language description to the solver program.


\begin{figure}
  \centering
  \includegraphics[width=0.44\textwidth]{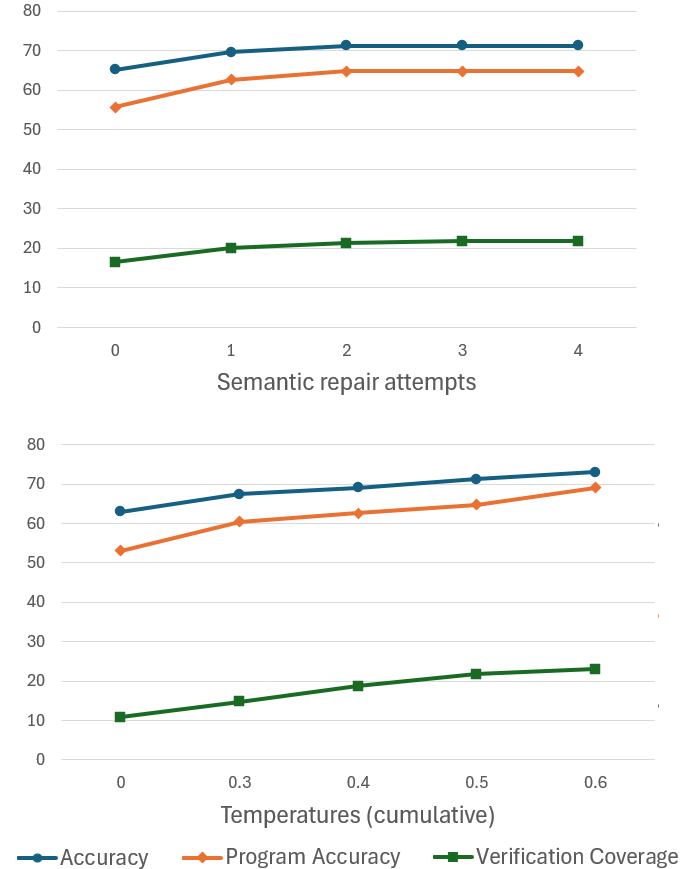} 
  \caption{Repair attempts and temperature variations on AR-LSAT}
  \label{fig:repair-temperature-variations}
\end{figure}

2. \underline{SSV verification has perfect precision across all datasets.} With GPT-4 as base model, SSV achieves 100\% verification precision on all datasets. Notably, on AR-LSAT, FOLIO, and ProofWriter, our verification mechanism identified erroneous cases where the datasets contained incorrect answers. However, for comparison with baselines, in Table \ref{tab:eval-general} we also report results based on the original datasets (showing slightly lower precision due to mislabelled cases). See the appendix for details of corrections. For AR-LSAT cases we also verified our corrections against the original test answers\footnote{\href{https://img.cracklsat.net/lsat/pt/pt80.pdf}{https://img.cracklsat.net/lsat/pt/pt80.pdf}}. This empirically perfect precision highlights SSV's robustness on complex reasoning tasks.

3. \underline{SSV verification has significant coverage on all datasets.} Although the precision is very high, we know that SSV verification does not always succeed. However, we find that the coverage is  significant across all datasets, with the lowest coverage of 21.7\% on the most difficult AR-LSAT dataset. As expected, we find the coverage increases on the relatively easier datasets, with a verification coverage of up to 75.2\% on ProofWriter. This significant coverage of verification shows that the SSV approach can help in avoiding manual human verification in a significant proportion of cases to reduce overall cost and effort.   


{\bf Effect of semantic repair and temperature exploration.} Figure \ref{fig:repair-temperature-variations} shows the impact of varying semantic repair attempts ($\mathtt{MaxRepairs}$) and temperatures ($\mathtt{Temperatures}$) on the AR-LSAT dataset. We analyze overall accuracy, program accuracy (how often program generation succeeds rather than direct LLM answers), and verification coverage. Semantic repair improves accuracy by 6.1\%, while temperature exploration increases it by 10.0\%. Verification coverage gains 5.2\% with repair and more than doubles with temperature exploration, rising 12.2\% above an initial 10.9\%. Repair attempts yield diminishing returns and cease to improve any metric beyond three attempts, while temperature exploration continues to show some gains up to 0.6. Additionally, the gap between program accuracy and overall accuracy narrows (from 9.8\% to 5.2\%, when averaged over both temperature and repair attempts), indicating greater reliance on program generation with these enhancements.

We also ran a full ablation on AR-LSAT without any repair or temperature sampling (effectively replicating Logic-LM but using compositional code generation). This scored 55.7\% vs. our 71.3\% (Logic-LM: 43\%), showing our novel features add 15.6\%, and other enhancements  contribute 12.7\%.

{\bf Evaluation on GPT-3.5.} We also evaluated our system and all baselines using GPT-3.5 as the underlying LLM. The results are shown in Table \ref{tab:eval-gpt35}. Firstly, we note that while the general accuracy of all systems drops significantly with this weaker model, our SSV system still performs best overall, with an average accuracy of 56.2\%. However, Logic-LM performs better than SSV on FOLIO and LogicalDeduction (this could be partly due to differences in the code generation quality for the different solver languages that Logic-LM uses for these datasets). Secondly, we observe that while the coverage of SSV verification also drops significantly, with two of the more difficult datasets (AR-LSAT and LogicalDeduction) having no coverage at all, the precision of SSV is very minimally affected. On the three datasets where there is coverage, we still see an average precision of 97\%. This demonstrates an important property of reliability of SSV verification: even for weaker models, if verification succeeds then it is still very reliable (and much more reliable than general accuracy), though it may succeed much less often. In practical terms, such reliability could even allow one to adopt a tiered strategy to optimize costs: trying weaker (cheaper) models for tasks first and fall-back on more expensive models if verification fails.


{\bf Verification failures} We  conducted a manual analysis on a sample of cases where verification did not pass. Classification of key reasons:  program not well-formed (13.3\%), program incorrect (53.3\%), example incorrect (10\%), both incorrect (23.3\%). Thus in most cases the program was incorrect, which aligns with the expectation that examples inference is generally simpler than abstract program formulation.

\begin{table*}[t]
\centering
\small{
\begin{tabular}{c||c|c|c|c||c|c}
\hline
\rule{0pt}{2.5ex}
\multirow{2}{*}{\textbf{Dataset}} & \multicolumn{4}{c||}{\textbf{General Accuracy}} & \multicolumn{2}{c}{\textbf{SSV Verification}} \\
\cline{2-7} 
 & \rule{0pt}{2.5ex} \textbf{Standard} & \textbf{\quad CoT\quad} & \textbf{Logic-LM} & \textbf{\quad SSV\quad} & \textbf{Coverage} & \textbf{Precision} \\
\hline \rule{0pt}{2.5ex}
AR-LSAT  & 20.3    & 17.3    & 26.4 & \bf{28.3} & 0 &  - \\ \rule{0pt}{2.5ex} 
FOLIO  &  45.1   & 57.4   & \bf{62.7} & 59.3 & 1.5 & 100.0 \\ \rule{0pt}{2.5ex}
LogicalDeduction & 40.0  & 42.3 & \bf{65.7}  & 48.3 & 0 & -  \\ \rule{0pt}{2.5ex} 
PrOntoQA & 47.4  &  67.8  & 61.0 & \bf{72.8} & 4.2 & 95.2 \\ \rule{0pt}{2.5ex} 
ProofWriter & 35.5  &  49.2  & 58.3 & \bf{72.5} & 16.2  & 94.8 \textcolor{darkgreen}{(95.9)} \\ \hline
\end{tabular}
}

\caption{General accuracy and SSV precision/coverage with GPT-3.5 base model. \emph{Values in brackets are actual values on corrected datasets.}}
\label{tab:eval-gpt35}
\end{table*}

\section{Limitations and Future Directions}
\label{sec:relatedwork}

Since natural language is informal, any verification approach with NL specifications cannot guarantee full correctness. While SSV verification achieves near-perfect empirical precision (100\% with GPT-4), we discuss the kinds of errors illustrated by some cases of incorrect verification observed with GPT-3.5 (specifically, one case in PrOntoQA and four in ProofWriter where incorrect answers passed verification).

1. \emph{Concrete instantiations are insufficient}. Since verification relies on concrete examples (test cases), these may not cover all aspects of a general constraint, particularly corner cases. This caused two failures with GPT-3.5. For instance, in one case, the conditions ``Gary is nice" and ``Gary is kind" were conflated into a single predicate ``is\_kind(Gary)" in the formalization. An instantiation asserting ``Gary is nice but not kind" could have detected this error.

2. \emph{Concrete instantiation and program are both mutually consistent but wrong}. This is the unlikely case where both the program and the test case have the same error and therefore pass verification. We found only one such case which was a rather confusingly trivial error: for some reason the constraint ``Fiona is quiet" was translated as its negation ``Not(is\_quiet(Fiona))" in both the program and the concrete instantiation independently generated by GPT-3.5. 

3. \emph{Missing or superfluous constraints}. 
The LLM may omit required constraints or introduce unintended ones. Since our approach relies on explicitly demarcated constraints parsed from the LLM-generated program, such errors can cause verification failures. Two GPT-3.5 failures resulted from superfluous constraints.

In general, such errors are rare, more common in weaker LLMs, and expected to decrease as LLMs improve. Errors of types (1) and (2) could be mitigated with a more exhaustive examples inference strategy, as our implementation generates only one positive and one negative example per constraint. Class (3) errors arise from structural inconsistencies where program constraints do not match the original problem. Such cases may be addressed by training specialized modules to more robustly enforce core structural properties.  

Another potential limitation is that while industrial provers like Z3 are effectively decidable for many practical problems (we observed no failures due to the solver), in more complex cases our method will conservatively fail verification, as decidability of first-order logic is undecidable in general. Future work may also explore addressing this limitation using iterative LLM reasoning to assist solver convergence.

\section{Related Work}
\label{sec:relatedwork}

{\bf Reasoning with LLMs}. Improving the robustness of reasoning in large language models is a very active area of research. One direction of work is to fine-tune or train specialized models that show improved reasoning ability \cite{entailer,transformers-soft-reasoners,nl-proofs-verifier-guided}. Another direction is to develop sophisticated prompting strategies to elicit better reasoning from LLMs. Chain-of-thought prompting \cite{cot} has shown how the quality of reasoning can be improved by prompting the model to explicitly generate the steps of reasoning in natural language before arriving at the final answer. Other examples of prompting approaches include self-consistency \cite{self-consistency}, analogical reasoning  \cite{thought-propagation}, and various modular approaches to address complex problems by decomposition to simpler sub-problems \cite{least-to-most,decomposed-prompting,selection-inference}. While these approaches show relative improvements in accuracy, the reasoning is still based on informal natural language and is prone to errors in the reasoning steps. In contrast, we follow the approach of off-loading the reasoning task to a formal solver that can guarantee correctness of the reasoning steps. Our particular focus is on the key challenge of ensuring  correct formalization of the problem.

{\bf Tool-augmented reasoning.} Integrating LLMs with specialized tools for performing various tasks is becoming increasingly common \cite{toolformer}. This approach has also been adopted to improve the reasoning quality by augmenting the LLM with logical solvers or automated reasoning tools \cite{logic-lm,sat-lm,dual-system-neurosymbolic}. The key challenge with these approaches is to ensure that the LLM correctly translates the reasoning problem from NL to the formal language of the solver. This is the main focus of our work, where we show how verification and refinement with respect to concrete instantiations generated by the LLM can both improve  accuracy and also provide verification with near-perfect precision. \cite{llmarc} also infer logic programs with test cases, but their test cases are arbitrary logical expressions inferred together with the program, and thus prone to similar errors the LLM may make in the program. In contrast, we generate concrete instantiations (literal assignments) independently from the program constraints, which the LLM can infer from the NL without any logical formulation. This yields very high precision verification which we can offer as a standalone feature, unlike any prior work. Tool-augmented approaches have also been explored in the related areas of planning \cite{planning-llm-modulo,planning-llms} and auto-formalization  \cite{autoformalization-llms,draft-sketch-prove,word-problems-llms-solvers}, where informal mathematical proofs are translated to formal specifications defined in theorem provers like Isabelle \cite{isabelle} and Lean \cite{lean}. While our work focuses on logical reasoning, the principle of consistency-based verificaion and refinement of formalizations using concrete instantiations is also potentially applicable to these other domains.

{\bf Self-verification approaches.} Many related works have also explored the notion of self-verification by LLMs  \cite{self-verification-emnlp23,self-refine,self-evaluation,deductive-verification-cot,self-check}. The general idea is that using the LLM to inspect and verify its own reasoning can show improvements, though in some domains self-critiquing has also shown diminished performance \cite{self-verification-planning-limitations}. Our approach of verification is different: instead of asking the LLM to verify the abstract chain of reasoning, we only ask it to generate concrete examples of the general constraints in the problem. The task of verification is then done with the solver to formally check that the examples are consistent with the abstract formalization. Thus apart from not relying purely on the LLM for verification, we also avoid the more complex task of verifying an abstract chain of reasoning which can itself be highly error-prone. We show how this approach provides a very high precision verification, as opposed to just relative improvements in accuracy. 




\section{Conclusion}
\label{sec:conclusion}

We have presented the Semantic Self-Verification approach, which infers strong problem formalizations based on concrete instantiations, using a consistency-based verification paradigm that leverages LLMs and logical solvers. Beyond achieving state-of-the-art accuracy, SSV introduces a novel verification feature that has near-perfect empirical precision. As the reasoning power of LLMs continues to advance, such near-certain verification can serve as a complementary dimension to general accuracy gains in order to ensure confidence on arbitrarily complex tasks. 


\paragraph{Acknowledgments.} The first author is grateful for the discussions with his daughter to help with her middle school studies, which provided the inspiration for this work.

\bibliographystyle{named}
\bibliography{ssv}


\section{Appendix}
\label{sec:appendix}

\subsection{Compositional Code Generation and Refinement Prompts}
\label{app:codegen}

\subsubsection{Problem Decomposition Prompt}

\texttt{Given a problem description, please decompose it into an initial context and a list of independent constraints. If there is no explicit initial context given and only constraints are given, then just state "None" for initial context. Some examples are given below.} \\
\texttt{------} \\
\texttt{\bf Problem:} \\
\texttt{The bald eagle eats the cow. The bald eagle is red. The bald eagle needs the cow. The bear needs the rabbit. The cow is kind. The cow is red. The cow needs the bald eagle. The rabbit eats the bear. The rabbit eats the cow. The rabbit sees the cow. If something needs the bald eagle then it needs the rabbit. If the bald eagle is nice and the bald eagle is young then the bald eagle sees the cow. If the rabbit needs the cow then the cow sees the rabbit. If something eats the cow and the cow is nice then it needs the bald eagle. If something needs the rabbit then it is nice. If something sees the rabbit then it is red. If something needs the bald eagle then it eats the bald eagle.} \\
\texttt{\bf InitialContext:} \\
\texttt{None} \\
\texttt{\bf Constraints:} \\
\texttt{The bald eagle eats the cow.} \\
\texttt{\#\#\#} \\
\texttt{The bald eagle is red.} \\
\texttt{\#\#\#} \\
\texttt{The bald eagle needs the cow.} \\
\texttt{\#\#\#} \\
\texttt{The bear needs the rabbit.} \\
\texttt{\#\#\#} \\
\texttt{The cow is kind.} \\
\texttt{\#\#\#} \\
\texttt{The cow is red.} \\
\texttt{\#\#\#} \\
\texttt{The cow needs the bald eagle.} \\
\texttt{\#\#\#} \\
\texttt{The rabbit eats the bear.} \\
\texttt{\#\#\#} \\
\texttt{The rabbit eats the cow.} \\
\texttt{\#\#\#} \\
\texttt{The rabbit sees the cow.} \\
\texttt{\#\#\#} \\
\texttt{If something needs the bald eagle then it needs the rabbit.} \\
\texttt{\#\#\#} \\
\texttt{If the bald eagle is nice and the bald eagle is young then the bald eagle sees the cow.} \\
\texttt{\#\#\#} \\
\texttt{If the rabbit needs the cow then the cow sees the rabbit.} \\
\texttt{\#\#\#} \\
\texttt{If something eats the cow and the cow is nice then it needs the bald eagle.} \\
\texttt{\#\#\#} \\
\texttt{If something needs the rabbit then it is nice.} \\
\texttt{\#\#\#} \\
\texttt{If something sees the rabbit then it is red.} \\
\texttt{\#\#\#} \\
\texttt{If something needs the bald eagle then it eats the bald eagle.} \\
\texttt{------} \\
\texttt{\bf Problem:} \\
\texttt{On Tuesday Vladimir and Wendy each eat exactly four separate meals: breakfast, lunch, dinner, and a snack. The following is all that is known about what they eat during that day: At no meal does Vladimir eat the same kind of food as Wendy. Neither of them eats the same kind of food more than once during the day. For breakfast, each eats exactly one of the following: hot cakes, poached eggs, or omelet. For lunch, each eats exactly one of the following: fish, hot cakes, macaroni, or omelet. For dinner, each eats exactly one of the following: fish, hot cakes, macaroni, or omelet. For a snack, each eats exactly one of the following: fish or omelet. Wendy eats an omelet for lunch.} \\
\texttt{\bf InitialContext:} \\
\texttt{On Tuesday Vladimir and Wendy each eat exactly four separate meals: breakfast, lunch, dinner, and a snack.} \\
\texttt{\bf Constraints:} \\
\texttt{At no meal does Vladimir eat the same kind of food as Wendy.} \\
\texttt{\#\#\#} \\
\texttt{Neither of them eats the same kind of food more than once during the day.} \\
\texttt{\#\#\#} \\
\texttt{For breakfast, each eats exactly one of the following: hot cakes, poached eggs, or omelet.} \\
\texttt{\#\#\#} \\
\texttt{For lunch, each eats exactly one of the following: fish, hot cakes, macaroni, or omelet.} \\
\texttt{\#\#\#} \\
\texttt{For dinner, each eats exactly one of the following: fish, hot cakes, macaroni, or omelet.} \\
\texttt{\#\#\#} \\
\texttt{For a snack, each eats exactly one of the following: fish or omelet.} \\
\texttt{\#\#\#} \\
\texttt{Wendy eats an omelet for lunch.} \\
\texttt{------} \\
\texttt{\bf Problem:} \\
\texttt{In a repair facility there are exactly six technicians: Stacy, Urma, Wim, Xena, Yolanda, and Zane. Each technician repairs machines of at least one of the following three types-radios, televisions, and VCRs-and no other types. The following conditions apply: Xena and exactly three other technicians repair radios. Yolanda repairs both televisions and VCRs. Stacy does not repair any type of machine that Yolanda repairs. Zane repairs more types of machines than Yolanda repairs. Wim does not repair any type of machine that Stacy repairs. Urma repairs exactly two types of machines.} \\
\texttt{\bf InitialContext:} \\
\texttt{In a repair facility there are exactly six technicians: Stacy, Urma, Wim, Xena, Yolanda, and Zane. Each technician repairs machines of at least one of the following three types-radios, televisions, and VCRs-and no other types.} \\
\texttt{\bf Constraints:} \\
\texttt{Xena and exactly three other technicians repair radios.} \\
\texttt{\#\#\#} \\
\texttt{Yolanda repairs both televisions and VCRs.} \\
\texttt{\#\#\#} \\
\texttt{Stacy does not repair any type of machine that Yolanda repairs.} \\
\texttt{\#\#\#} \\
\texttt{Zane repairs more types of machines than Yolanda repairs.} \\
\texttt{\#\#\#} \\
\texttt{Wim does not repair any type of machine that Stacy repairs.} \\
\texttt{\#\#\#} \\
\texttt{Urma repairs exactly two types of machines.} \\
\texttt{------} \\

\subsubsection{Incremental Code Generation Prompt}

\texttt{Given a z3 program that models a particular problem and a new constraint described in natural language, please provide the z3 code to augment the program with the new constraint. Please provide only the z3 program code in the output and no other markdown formatting or explanatory text.} \\
\texttt{------} \\
\texttt{\bf ExistingProgram:} \\
\texttt{\# On Tuesday Vladimir and Wendy each eat exactly four separate meals: breakfast, lunch, dinner, and a snack.} \\
\texttt{from z3 import *} \\
\texttt{people\_sort, (Vladimir, Wendy) = EnumSort('people', ['Vladimir', 'Wendy'])} \\
\texttt{meals\_sort, (breakfast, lunch, dinner, snack) = EnumSort('meals', ['breakfast', 'lunch', 'dinner', 'snack'])} \\
\texttt{foods\_sort, (fish, hot\_cakes, macaroni, omelet, poached\_eggs) = EnumSort('foods', ['fish', 'hot\_cakes', 'macaroni', 'omelet', 'poached\_eggs'])} \\
\texttt{people = [Vladimir, Wendy]} \\
\texttt{meals = [breakfast, lunch, dinner, snack]} \\
\texttt{foods = [fish, hot\_cakes, macaroni, omelet, poached\_eggs]} \\
\texttt{eats = Function('eats', people\_sort, meals\_sort, foods\_sort)} \\

\texttt{pre\_conditions = []} \\

\texttt{\# CONSTRAINT: At no meal does Vladimir eat the same kind of food as Wendy.} \\
\texttt{m = Const('m', meals\_sort)} \\
\texttt{pre\_conditions.append(ForAll([m], eats(Vladimir, m) != eats(Wendy, m)))} \\

\texttt{\bf NewConstraint:} \\
\texttt{Neither of them eats the same kind of food more than once during the day.} \\
\texttt{\bf NewConstraintCode:} \\
\texttt{m = Const('m', meals\_sort)} \\
\texttt{p = Const('p', people\_sort)} \\
\texttt{f = Const('f', foods\_sort)} \\
\texttt{pre\_conditions.append(ForAll([p, f], Sum([eats(p, m) == f for m in meals]) <= 1))} \\
\texttt{------} \\
\texttt{\bf ExistingProgram:} \\
\texttt{\# In a repair facility there are exactly six technicians: Stacy, Urma, Wim, Xena, Yolanda, and Zane. Each technician repairs machines of at least one of the following three types—radios, televisions, and VCRs—and no other types.} \\
\texttt{from z3 import *} \\
\texttt{technicians\_sort, (Stacy, Urma, Wim, Xena, Yolanda, Zane) = EnumSort('technicians', ['Stacy', 'Urma', 'Wim', 'Xena', 'Yolanda', 'Zane'])} \\
\texttt{machines\_sort, (radios, televisions, VCRs) = EnumSort('machines', ['radios', 'televisions', 'VCRs'])} \\
\texttt{technicians = [Stacy, Urma, Wim, Xena, Yolanda, Zane]} \\
\texttt{machines = [radios, televisions, VCRs]} \\
\texttt{repairs = Function('repairs', technicians\_sort, machines\_sort, BoolSort())} \\

\texttt{pre\_conditions = []} \\
\texttt{t = Const('t', technicians\_sort)} \\
\texttt{pre\_conditions.append(ForAll([t], Sum([repairs(t, m) for m in machines]) >= 1))} \\

\texttt{\bf NewConstraint:} \\
\texttt{Xena and exactly three other technicians repair radios.} \\
\texttt{\bf NewConstraintCode:} \\
\texttt{t = Const('t', technicians\_sort)} \\
\texttt{pre\_conditions.append(And(repairs(Xena, radios), Sum([And(t != Xena, repairs(t, radios)) for t in technicians]) == 3))} \\
\texttt{------} \\

\subsubsection{Options Code Generation Prompt}

\texttt{Given a problem with multiple answer options and an existing z3 program that models the problem, please provide the z3 code that checks each option and prints the correct answer. For each option, first create the check\_property for the option by substituting the option values appropriately in the question statement, as well as a full comment describing what the check\_property is stating. Then use only the following custom functions (is\_unsat(), is\_sat() and is\_valid()) to check if the check\_property is unsatisfiable, satisfiable or valid (depending on the question). Please structure the code with comments exactly as shown in the few shot examples below. Please provide only the options code and its comments in the output (not the full program), and no other surrounding markdown formatting or explanatory text. Please create independently executable code for each option (even if the option is not satisfiable) and do not share code between different options.} \\

\texttt{def is\_unsat(option\_constraints):} \\
\texttt{solver = Solver()} \\
\texttt{solver.add(pre\_conditions)} \\
\texttt{solver.add(option\_constraints)} \\
\texttt{return solver.check() == unsat} \\

\texttt{def is\_sat(option\_constraints):} \\
\texttt{solver = Solver()} \\
\texttt{solver.add(pre\_conditions)} \\
\texttt{return solver.check() == sat} \\

\texttt{def is\_valid(option\_constraints):} \\
\texttt{return is\_sat(option\_constraints) and is\_unsat(Not(option\_constraints))} \\
\texttt{------} \\
\texttt{>>> \bf Problem:} \\
\texttt{On Tuesday Vladimir and Wendy each eat exactly four separate meals: breakfast, lunch, dinner, and a snack.} \\
\texttt{>>> \bf ExistingProgram:} \\
\texttt{from z3 import *} \\
\texttt{people\_sort, (Vladimir, Wendy) = EnumSort('people', ['Vladimir', 'Wendy'])} \\
\texttt{meals\_sort, (breakfast, lunch, dinner, snack) = EnumSort('meals', ['breakfast', 'lunch', 'dinner', 'snack'])} \\
\texttt{foods\_sort, (fish, hot\_cakes, macaroni, omelet, poached\_eggs) = EnumSort('foods', ['fish', 'hot\_cakes', 'macaroni', 'omelet', 'poached\_eggs'])} \\
\texttt{people = [Vladimir, Wendy]} \\
\texttt{meals = [breakfast, lunch, dinner, snack]} \\
\texttt{foods = [fish, hot\_cakes, macaroni, omelet, poached\_eggs]} \\
\texttt{eats = Function('eats', people\_sort, meals\_sort, foods\_sort)} \\

\texttt{pre\_conditions = []} \\

\texttt{\# CONSTRAINT: At no meal does Vladimir eat the same kind of food as Wendy.} \\
\texttt{m = Const('m', meals\_sort)} \\
\texttt{pre\_conditions.append(ForAll([m], eats(Vladimir, m) != eats(Wendy, m)))} \\

\texttt{\# CONSTRAINT: Neither of them eats the same kind of food more than once during the day.} \\
\texttt{m = Const('m', meals\_sort)} \\
\texttt{p = Const('p', people\_sort)} \\
\texttt{f = Const('f', foods\_sort)} \\
\texttt{pre\_conditions.append(ForAll([p, f], Sum([eats(p, m) == f for m in meals]) <= 1))} \\

\texttt{>>> \bf Question:} \\
\texttt{Vladimir cannot eat which one of the following foods?} \\
\texttt{>>> \bf Options:} \\
\texttt{(A) fish} \\
\texttt{(B) hot cakes} \\
\texttt{(C) macaroni} \\
\texttt{(D) omelet} \\
\texttt{(E) poached eggs} \\
\texttt{>>> \bf OptionsCode:} \\

\texttt{\# CHECK TYPE: question says "cannot" so will check for validity using is\_valid() to ensure that the negated statement is true in all possible models.} \\

\texttt{\# OPTION A:} \\
\texttt{\# CHECK PROPERTY: Vladimir cannot eat which one of the following foods? ANSWER: fish.} \\
\texttt{m = Const('m', meals\_sort)} \\
\texttt{check\_property = ForAll([m], eats(Vladimir, m) != fish)} \\
\texttt{if is\_valid(check\_property): print('(A)')} \\

\texttt{\# OPTION B:} \\
\texttt{\# CHECK PROPERTY: Vladimir cannot eat which one of the following foods? ANSWER: hot cakes.} \\
\texttt{m = Const('m', meals\_sort)} \\
\texttt{check\_property = ForAll([m], eats(Vladimir, m) != hot\_cakes)} \\
\texttt{if is\_valid(check\_property): print('(B)')} \\

\texttt{\# OPTION C:} \\
\texttt{\# CHECK PROPERTY: Vladimir cannot eat which one of the following foods? ANSWER: macaroni.} \\
\texttt{m = Const('m', meals\_sort)} \\
\texttt{check\_property = ForAll([m], eats(Vladimir, m) != macaroni)} \\
\texttt{if is\_valid(check\_property): print('(C)')} \\

\texttt{\# OPTION D:} \\
\texttt{\# CHECK PROPERTY: Vladimir cannot eat which one of the following foods? ANSWER: omelet.} \\
\texttt{m = Const('m', meals\_sort)} \\
\texttt{check\_property = ForAll([m], eats(Vladimir, m) != omelet)} \\
\texttt{if is\_valid(check\_property): print('(D)')} \\

\texttt{\# OPTION E:} \\
\texttt{\# CHECK PROPERTY: Vladimir cannot eat which one of the following foods? ANSWER: poached eggs.} \\
\texttt{m = Const('m', meals\_sort)} \\
\texttt{check\_property = ForAll([m], eats(Vladimir, m) != poached\_eggs)} \\
\texttt{if is\_valid(check\_property): print('(E)')} \\

\texttt{------} \\
\texttt{>>> \bf Problem:} \\
\texttt{In a repair facility there are exactly six technicians: Stacy, Urma, Wim, Xena, Yolanda, and Zane. Each technician repairs equipment of at least one of the following three types—radios, televisions, and VCRs—and no other types.} \\
\texttt{>>> \bf ExistingProgram:} \\
\texttt{from z3 import *} \\
\texttt{technicians\_sort, (Stacy, Urma, Wim, Xena, Yolanda, Zane) = EnumSort('technicians', ['Stacy', 'Urma', 'Wim', 'Xena', 'Yolanda', 'Zane'])} \\
\texttt{equipment\_sort, (radios, televisions, VCRs) = EnumSort('equipment', ['radios', 'televisions', 'VCRs'])} \\
\texttt{technicians = [Stacy, Urma, Wim, Xena, Yolanda, Zane]} \\
\texttt{equipment = [radios, televisions, VCRs]} \\
\texttt{repairs = Function('repairs', technicians\_sort, equipment\_sort, BoolSort())} \\

\texttt{pre\_conditions = []} \\
\texttt{t = Const('t', technicians\_sort)} \\
\texttt{pre\_conditions.append(ForAll([t], Sum([repairs(t, e) for e in equipment]) >= 1))} \\

\texttt{\# CONSTRAINT: Xena and exactly three other technicians repair radios.} \\
\texttt{t = Const('t', technicians\_sort)} \\
\texttt{pre\_conditions.append(And(repairs(Xena, radios), Sum([And(t != Xena, repairs(t, radios)) for t in technicians]) == 3))} \\

\texttt{>>> \bf Question:} \\
\texttt{Which one of the following can be a complete and accurate list of the technicians that repair televisions?} \\
\texttt{>>> \bf Options:} \\
\texttt{(A) Stacy, Wim, Zane} \\
\texttt{(B) Urma, Wim, Xena, Yolanda} \\
\texttt{(C) Xena, Yolanda} \\
\texttt{(D) Stacy, Urma, Wim, Xena, Yolanda, Zane} \\
\texttt{(E) Urma} \\
\texttt{>>> \bf OptionsCode:} \\

\texttt{\# CHECK TYPE: question says "can be" so will check for satisfiable using is\_sat()} \\

\texttt{\# OPTION A:} \\
\texttt{\# CHECK PROPERTY: Which one of the following can be a complete and accurate list of the technicians that repair televisions? ANSWER: Stacy, Wim and Zane.} \\
\texttt{e = Const('e', equipment\_sort)} \\
\texttt{check\_property = And(repairs(Stacy, televisions), repairs(Wim, televisions), repairs(Wim, televisions), Not(repairs(Urma, televisions)), Not(repairs(Xena, televisions)), Not(repairs(Yolanda, televisions)))} \\
\texttt{if is\_sat(check\_property): print('(A)')} \\

\texttt{\# OPTION B:} \\
\texttt{\# CHECK PROPERTY: Which one of the following can be a complete and accurate list of the technicians that repair televisions? ANSWER: Urma, Wim, Xena and Yolanda.} \\
\texttt{e = Const('e', equipment\_sort)} \\
\texttt{check\_property = And(repairs(Urma, televisions), repairs(Wim, televisions), repairs(Xena, televisions), repairs(Yolanda, televisions), Not(repairs(Stacy, televisions)), Not(repairs(Zane, televisions)))} \\
\texttt{if is\_sat(check\_property): print('(B)')} \\

\texttt{\# OPTION C:} \\
\texttt{\# CHECK PROPERTY: Which one of the following can be a complete and accurate list of the technicians that repair televisions? ANSWER: Xena and Yolanda.} \\
\texttt{e = Const('e', equipment\_sort)} \\
\texttt{check\_property = And(repairs(Xena, televisions), repairs(Yolanda, televisions), Not(repairs(Stacy, televisions)), Not(repairs(Urma, televisions)), Not(repairs(Wim, televisions)), Not(repairs(Zane, televisions)))} \\
\texttt{if is\_sat(check\_property): print('(C)')} \\

\texttt{\# OPTION D:} \\
\texttt{\# CHECK PROPERTY: Which one of the following can be a complete and accurate list of the technicians that repair televisions? ANSWER: Stacy, Urma, Wim, Xena, Yolanda and Zane.} \\
\texttt{e = Const('e', equipment\_sort)} \\
\texttt{check\_property = And(repairs(Stacy, televisions), repairs(Urma, televisions), repairs(Wim, televisions), repairs(Xena, televisions), repairs(Yolanda, televisions), repairs(Zane, televisions))} \\
\texttt{if is\_sat(check\_property): print('(D)')} \\

\texttt{\# OPTION E:} \\
\texttt{\# CHECK PROPERTY: Which one of the following can be a complete and accurate list of the technicians that repair televisions? ANSWER: Urma.} \\
\texttt{e = Const('e', equipment\_sort)} \\
\texttt{check\_property = And(repairs(Urma, televisions), Not(repairs(Stacy, televisions)), Not(repairs(Wim, televisions)), Not(repairs(Xena, televisions)), Not(repairs(Yolanda, televisions)), Not(repairs(Zane, televisions)))} \\
\texttt{if is\_sat(check\_property): print('(E)')} \\
\texttt{------} \\

\subsubsection{Error-based Code Refinement Prompt}

\texttt{We are given a z3 program and an error message obtained from running it. First, please provide an analysis that investigates what may be the problem in the program that may be causing the error. Then, based on this analysis, please provide the corrected program where the issue is fixed - please make sure to retain any comments from the original code in the repaired code (especially the "CONSTRAINT", "QUESTION" or "OPTION" comments which demarcate special code segments - please do not remove, change or add any new such comments). If there is a general issue in the formulation, then please consider an alternative reformulation so that the program can execute without errors. A couple of sample cases are shown below for illustration. Please produce output in exactly the format shown in these samples, with the ">>> CorrectedProgram:" label clearly demarcating the corrected code, and do not use any other markdown formatting.} \\
\texttt{------} \\
\texttt{>>> \bf ExistingProgram:} \\
\texttt{from z3 import *} \\
\texttt{people\_sort, (Vladimir, Wendy) = EnumSort('people', ['Vladimir', 'Wendy'])} \\
\texttt{meals\_sort, (breakfast, lunch, dinner, snack) = EnumSort('meals', ['breakfast', 'lunch', 'dinner', 'snack'])} \\
\texttt{foods\_sort, (fish, hot\_cakes, macaroni, omelet, poached\_eggs) = EnumSort('foods', ['fish', 'hot\_cakes', 'macaroni', 'omelet', 'poached\_eggs'])} \\
\texttt{people = [Vladimir, Wendy]} \\
\texttt{meals = [breakfast, lunch, dinner, snack]} \\
\texttt{foods = [fish, hot\_cakes, macaroni, omelet, poached\_eggs]} \\
\texttt{eats = Function('eats', people\_sort, meals\_sort, foods\_sort)} \\

\texttt{pre\_conditions = []} \\

\texttt{\# CONSTRAINT: At no meal does Vladimir eat the same kind of food as Wendy.} \\
\texttt{pre\_conditions.append(ForAll([m], eats(Vladimir, m) != eats(Wendy, m)))} \\
\texttt{>>> \bf ErrorMessage:} \\
\texttt{"NameError: name 'm' is not defined"} \\
\texttt{>>> \bf ProblemDiscussion:} \\
\texttt{This program defines three enumerations for people, meals, and foods. It then specifies that Vladimir and Wendy are people, and lists the available meals and foods. It also creates a function eats which represents the food each person eats at each meal. Finally, it tries to add a constraint to ensure that Vladimir and Wendy do not eat the same kind of food at any meal. However, as the error message indicates, the constraint code uses a variable 'm' that has not been previously declared. Hence the correct fix to this issue would be to first explicitly declare the variable 'm' as a new const of meal\_sort.} \\
\texttt{>>> \bf CorrectedProgram:} \\
\texttt{from z3 import *} \\
\texttt{people\_sort, (Vladimir, Wendy) = EnumSort('people', ['Vladimir', 'Wendy'])} \\
\texttt{meals\_sort, (breakfast, lunch, dinner, snack) = EnumSort('meals', ['breakfast', 'lunch', 'dinner', 'snack'])} \\
\texttt{foods\_sort, (fish, hot\_cakes, macaroni, omelet, poached\_eggs) = EnumSort('foods', ['fish', 'hot\_cakes', 'macaroni', 'omelet', 'poached\_eggs'])} \\
\texttt{people = [Vladimir, Wendy]} \\
\texttt{meals = [breakfast, lunch, dinner, snack]} \\
\texttt{foods = [fish, hot\_cakes, macaroni, omelet, poached\_eggs]} \\
\texttt{eats = Function('eats', people\_sort, meals\_sort, foods\_sort)} \\

\texttt{pre\_conditions = []} \\

\texttt{\# CONSTRAINT: At no meal does Vladimir eat the same kind of food as Wendy.} \\
\texttt{m = Const('m', meals\_sort)} \\
\texttt{pre\_conditions.append(ForAll([m], eats(Vladimir, m) != eats(Wendy, m)))} \\
\texttt{------} \\

\subsection{Instantiation generation prompt}
\label{app:instantiation}

\texttt{Given a problem scenario, some Z3 initialization code that defines the data structures, and a list of constraints, please provide positive and negative examples for each constraint. Each positive example should have a description and an expression of concrete assignments that satisfy the constraint, while each negative example should have a description and an expression of concrete assignments that contradict the constraint. If a constraint or its examples cannot be expressed by the given data structures or definitions, then please state "NONE" for the example description and "pass" for the assignments code. Please provide the completion to the prompt in exactly the same format as the example given below.} \\
\texttt{------} \\
\texttt{>>> \bf Scenario:} \\
\texttt{None} \\
\texttt{>>> \bf InitializationCode:} \\
\texttt{from z3 import *} \\
\texttt{creature\_sort = DeclareSort('creature')} \\
\texttt{Stella = Const('Stella', creature\_sort)} \\
\texttt{Jay = Const('Jay', creature\_sort)} \\
\texttt{is\_tumpus = Function('is\_tumpus', creature\_sort, BoolSort())} \\
\texttt{is\_rompus = Function('is\_rompus', creature\_sort, BoolSort())} \\
\texttt{is\_numpus = Function('is\_numpus', creature\_sort, BoolSort())} \\
\texttt{is\_yumpus = Function('is\_yumpus', creature\_sort, BoolSort())} \\
\texttt{is\_zumpus = Function('is\_zumpus', creature\_sort, BoolSort())} \\
\texttt{is\_impus = Function('is\_impus', creature\_sort, BoolSort())} \\
\texttt{is\_dumpus = Function('is\_dumpus', creature\_sort, BoolSort())} \\
\texttt{is\_vumpus = Function('is\_vumpus', creature\_sort, BoolSort())} \\
\texttt{is\_jompus = Function('is\_jompus', creature\_sort, BoolSort())} \\
\texttt{is\_wumpus = Function('is\_wumpus', creature\_sort, BoolSort())} \\
\texttt{is\_angry = Function('is\_angry', creature\_sort, BoolSort())} \\
\texttt{is\_bright = Function('is\_bright', creature\_sort, BoolSort())} \\
\texttt{is\_luminous = Function('is\_luminous', creature\_sort, BoolSort())} \\
\texttt{is\_transparent = Function('is\_transparent', creature\_sort, BoolSort())} \\
\texttt{is\_bitter = Function('is\_bitter', creature\_sort, BoolSort())} \\
\texttt{is\_red = Function('is\_red', creature\_sort, BoolSort())} \\
\texttt{is\_happy = Function('is\_happy', creature\_sort, BoolSort())} \\
\texttt{is\_large = Function('is\_large', creature\_sort, BoolSort())} \\

\texttt{pre\_conditions = []} \\
\texttt{>>> \bf Constraints:} \\
\texttt{Each dumpus is a vumpus.} \\
\texttt{\#\#\#} \\
\texttt{Vumpuses are bright.} \\
\texttt{\#\#\#} \\
\texttt{Every vumpus is a zumpus.} \\
\texttt{\#\#\#} \\
\texttt{Zumpuses are not luminous.} \\
\texttt{>>> \bf ConstraintExamples:} \\
\texttt{Constraint:} \\
\texttt{Each dumpus is a vumpus.} \\
\texttt{PositiveExampleDescription:} \\
\texttt{Stella is a dumpus and is also a vumpus.} \\
\texttt{PositiveExampleCode:} \\
\texttt{And(is\_dumpus(Stella) == True, is\_vumpus(Stella) == True)} \\
\texttt{NegativeExampleDescription:} \\
\texttt{Stella is a dumpus but is not a vumpus.} \\
\texttt{NegativeExampleCode:} \\
\texttt{And(is\_dumpus(Stella) == True, is\_vumpus(Stella) == False)} \\
\texttt{Constraint:} \\
\texttt{Vumpuses are bright.} \\
\texttt{PositiveExampleDescription:} \\
\texttt{Jay is a vumpus and is bright.} \\
\texttt{PositiveExampleCode:} \\
\texttt{And(is\_vumpus(Jay) == True, is\_bright(Jay) == True)} \\
\texttt{NegativeExampleDescription:} \\
\texttt{Jay is a vumpus and is not bright.} \\
\texttt{NegativeExampleCode:} \\
\texttt{And(is\_vumpus(Jay) == True, is\_bright(Jay) == False)} \\
\texttt{Constraint:} \\
\texttt{Every vumpus is a zumpus.} \\
\texttt{PositiveExampleDescription:} \\
\texttt{Jay is a vumpus and a zumpus.} \\
\texttt{PositiveExampleCode:} \\
\texttt{And(is\_vumpus(Jay) == True, is\_zumpus(Jay) == True)} \\
\texttt{NegativeExampleDescription:} \\
\texttt{Jay is a vumpus but not a zumpus.} \\
\texttt{NegativeExampleCode:} \\
\texttt{And(is\_vumpus(Jay) == True, is\_zumpus(Jay) == False)} \\
\texttt{Constraint:} \\
\texttt{Zumpuses are not luminous.} \\
\texttt{PositiveExampleDescription:} \\
\texttt{Stella is a zumpus and is not luminous.} \\
\texttt{PositiveExampleCode:} \\
\texttt{And(is\_zumpus(Stella) == True, is\_luminous(Stella) == False)} \\
\texttt{NegativeExampleDescription:} \\
\texttt{Stella is a zumpus and is luminous.} \\
\texttt{NegativeExampleCode:} \\
\texttt{And(is\_zumpus(Stella) == True, is\_luminous(Stella) == True)} \\
\texttt{------} \\
\texttt{>>> \bf Scenario:} \\
\texttt{On Tuesday Vladimir and Wendy each eat exactly two separate meals: breakfast and dinner.} \\
\texttt{>>> \bf InitializationCode:} \\
\texttt{from z3 import *} \\

\texttt{people\_sort, (Vladimir, Wendy) = EnumSort('people', ['Vladimir', 'Wendy'])} \\
\texttt{meals\_sort, (breakfast, dinner) = EnumSort('meals', ['breakfast', 'dinner'])} \\
\texttt{foods\_sort, (fish, hot\_cakes, macaroni, omelet, poached\_eggs) = EnumSort('foods', ['fish', 'hot\_cakes', 'macaroni', 'omelet', 'poached\_eggs'])} \\
\texttt{people = [Vladimir, Wendy]} \\
\texttt{meals = [breakfast, dinner]} \\
\texttt{foods = [fish, hot\_cakes, macaroni, omelet, poached\_eggs]} \\
\texttt{eats = Function('eats', people\_sort, meals\_sort, foods\_sort)} \\

\texttt{pre\_conditions = []} \\
\texttt{>>> \bf Constraints:} \\
\texttt{At no meal does Vladimir eat the same kind of food as Wendy.} \\
\texttt{\#\#\#} \\
\texttt{Neither of them eats the same kind of food more than once during the day.} \\
\texttt{\#\#\#} \\
\texttt{For breakfast, each eats hot cakes.} \\
\texttt{>>> \bf ConstraintExamples:} \\
\texttt{Constraint:} \\
\texttt{At no meal does Vladimir eat the same kind of food as Wendy.} \\
\texttt{PositiveExampleDescription:} \\
\texttt{Vladimir and Wendy eat different foods at each meal: Vladimir has fish for breakfast while Wendy has hot cakes, and for dinner, Vladimir eats macaroni while Wendy has omelet.} \\
\texttt{PositiveExampleCode:} \\
\texttt{And(eats(Vladimir, breakfast) == fish, eats(Wendy, breakfast) == hot\_cakes,} \\
\texttt{eats(Vladimir, dinner) == macaroni, eats(Wendy, dinner) == omelet)} \\
\texttt{NegativeExampleDescription:} \\
\texttt{At dinner, both Vladimir and Wendy eat the same food, macaroni.} \\
\texttt{NegativeExampleCode:} \\
\texttt{And(eats(Vladimir, dinner) == macaroni, eats(Wendy, dinner) == macaroni)} \\
\texttt{Constraint:} \\
\texttt{Neither of them eats the same kind of food more than once during the day.} \\
\texttt{PositiveExampleDescription:} \\
\texttt{Vladimir eats different foods for breakfast and dinner: fish for breakfast and hot cakes for dinner. Wendy also eats different foods for both meals: hot cakes for breakfast and omelet for dinner.} \\
\texttt{PositiveExampleCode:} \\
\texttt{And(eats(Vladimir, breakfast) == fish, eats(Vladimir, dinner) == hot\_cakes,} \\
\texttt{eats(Wendy, breakfast) == hot\_cakes, eats(Wendy, dinner) == omelet)} \\
\texttt{NegativeExampleDescription:} \\
\texttt{Vladimir eats fish for both breakfast and dinner.} \\
\texttt{NegativeExampleCode:} \\
\texttt{And(eats(Vladimir, breakfast) == fish, eats(Vladimir, dinner) == fish)} \\
\texttt{Constraint:} \\
\texttt{For breakfast, each eats hot cakes.} \\
\texttt{PositiveExampleDescription:} \\
\texttt{Vladimir and Wendy both eat hot cakes for breakfast.} \\
\texttt{PositiveExampleCode:} \\
\texttt{And(eats(Vladimir, breakfast) == hot\_cakes, eats(Wendy, breakfast) == hot\_cakes)} \\
\texttt{NegativeExampleDescription:} \\
\texttt{Vladimir eats macaroni for breakfast.} \\
\texttt{NegativeExampleCode:} \\
\texttt{eats(Vladimir, breakfast) == macaroni} \\
\texttt{------} \\
\texttt{>>> \bf Scenario:} \\
\texttt{In a repair facility there are exactly six technicians: Stacy, Urma, Wim, Xena, Yolanda, and Zane. Each technician repairs machines of at least one of the following three types—radios, televisions, and VCRs—and no other types.} \\
\texttt{>>> \bf InitializationCode:} \\
\texttt{from z3 import *} \\
\texttt{technicians\_sort, (Stacy, Urma, Wim, Xena, Yolanda, Zane) = EnumSort('technicians', ['Stacy', 'Urma', 'Wim', 'Xena', 'Yolanda', 'Zane'])} \\
\texttt{machines\_sort, (radios, televisions, VCRs) = EnumSort('machines', ['radios', 'televisions', 'VCRs'])} \\
\texttt{technicians = [Stacy, Urma, Wim, Xena, Yolanda, Zane]} \\
\texttt{machines = [radios, televisions, VCRs]} \\
\texttt{repairs = Function('repairs', technicians\_sort, machines\_sort, BoolSort())} \\

\texttt{pre\_conditions = []} \\
\texttt{t = Const('t', technicians\_sort)} \\
\texttt{pre\_conditions.append(ForAll([t], Sum([repairs(t, m) for m in machines]) >= 1))} \\

\texttt{>>> \bf Constraints:} \\
\texttt{Xena and exactly three other technicians repair radios.} \\
\texttt{\#\#\#} \\
\texttt{Stacy needs help repairing VCRs.} \\
\texttt{\#\#\#} \\
\texttt{Urma and Zane repair the same type of machine.} \\
\texttt{>>> \bf ConstraintExamples:} \\
\texttt{Constraint:} \\
\texttt{Xena and exactly three other technicians repair radios.} \\
\texttt{PositiveExampleDescription:} \\
\texttt{Only Xena, Wim, Yolanda, and Zane repair radios and no one else.} \\
\texttt{PositiveExampleCode:} \\
\texttt{And(repairs(Stacy, radios) == False, repairs(Urma, radios) == False, repairs(Wim, radios) == True, repairs(Xena, radios) == True, repairs(Yolanda, radios) == True, repairs(Zane, radios) == True)} \\
\texttt{NegativeExampleDescription:} \\
\texttt{Only Xena and Yolanda repair radios and no one else.} \\
\texttt{NegativeExampleCode:} \\
\texttt{And(repairs(Stacy, radios) == False, repairs(Urma, radios) == False, repairs(Wim, radios) == False, repairs(Xena, radios) == True, repairs(Yolanda, radios) == True, repairs(Zane, radios) == False)} \\
\texttt{Constraint:} \\
\texttt{Stacy needs help repairing VCRs.} \\
\texttt{PositiveExampleDescription:} \\
\texttt{NONE} \\
\texttt{PositiveExampleCode:} \\
\texttt{pass} \\
\texttt{NegativeExampleDescription:} \\
\texttt{NONE} \\
\texttt{NegativeExampleCode:} \\
\texttt{pass} \\
\texttt{Constraint:} \\
\texttt{Urma and Zane repair the same type of machine.} \\
\texttt{PositiveExampleDescription:} \\
\texttt{Urma and Zane both repair VCRs.} \\
\texttt{PositiveExampleCode:} \\
\texttt{And(repairs(Urma, VCRs) == True, repairs(Zane, VCRs) == True)} \\
\texttt{NegativeExampleDescription:} \\
\texttt{Urma repairs televisions, while Zane repairs radios.} \\
\texttt{NegativeExampleCode:} \\
\texttt{And(repairs(Urma, televisions) == True, repairs(Zane, radios) == True)} \\
\texttt{------} \\

\subsection{Semantic repair prompt}
\label{app:repair}

\texttt{We are given a scenario description, some initial z3 code that sets up basic definitions, a constraint in natural language, and a code snippet that implements that constraint. We are also given some code that should implement a positive example to the constraint, which should be satisfiable under that constraint, but it is not. First, please provide an analysis that investigates what may be the problem in either the initial code, the constraint code or the example. Then, based on this analysis, please repair the relevant code segments (initial code, constraint code, or example code) so that the positive example becomes satisfiable (state 'NONE' if no repair is required to a code segment). If multiple segments are incorrect due to a general formulation problem, then please reformulate the whole solution approach in the initial code and produce appropriate code for all segments. A couple of sample cases are shown below for illustration. Please produce output in exactly the format shown in these samples, and do not use any other markdown formatting.} \\
\texttt{------} \\
\texttt{\bf Scenario:} \\
\texttt{On Tuesday Vladimir and Wendy each eat exactly four separate meals: breakfast, lunch, dinner, and a snack.} \\
\texttt{\bf InitialCode:} \\
\texttt{from z3 import *} \\
\texttt{people\_sort, (Vladimir, Wendy) = EnumSort('people', ['Vladimir', 'Wendy'])} \\
\texttt{meals\_sort, (breakfast, lunch, dinner, snack) = EnumSort('meals', ['breakfast', 'lunch', 'dinner', 'snack'])} \\
\texttt{foods\_sort, (fish, hot\_cakes, macaroni, omelet, poached\_eggs) = EnumSort('foods', ['fish', 'hot\_cakes', 'macaroni', 'omelet', 'poached\_eggs'])} \\
\texttt{people = [Vladimir, Wendy]} \\
\texttt{meals = [breakfast, lunch, dinner, snack]} \\
\texttt{foods = [fish, hot\_cakes, macaroni, omelet, poached\_eggs]} \\
\texttt{eats = Function('eats', people\_sort, meals\_sort, foods\_sort)} \\

\texttt{pre\_conditions = []} \\

\texttt{\bf ConstraintDescription:} \\
\texttt{At some meal Vladimir eats the same kind of food as Wendy.} \\
\texttt{\bf ConstraintCode:} \\
\texttt{m = Const('m', meals\_sort)} \\
\texttt{pre\_conditions.append(ForAll([m], eats(Vladimir, m) != eats(Wendy, m)))} \\
\texttt{\bf PositiveExampleCode:} \\
\texttt{And(eats(Vladimir, breakfast) == fish, eats(Wendy, breakfast) == fish)} \\
\texttt{\bf ProblemDiscussion:} \\
\texttt{The scenario describes foods that Vladimir and Wendy eat at various meals during the day. The initial code defines the main data structures and the eats function which indicates the food each person eats on every meal. The constraint requires that there is at least one meal where they both eat the same food. The constraint code asserts that for all meals, the food that Vladimir eats is different from what Wendy eats. But this contradicts the intended constraint. The positive example code states that at breakfast, both Vladimir and Wendy eat fish, and this is consistent with the requirements of the constraint. Hence there is no issue in the initial code and the example code, but the constraint code wrongly implements the constraint. It should be repaired to assert that for some meal, both Vladimir and Wendy eat the same food.} \\
\texttt{\bf RepairedInitialCode:} \\
\texttt{NONE} \\
\texttt{\bf RepairedConstraintCode:} \\
\texttt{m = Const('m', meals\_sort)} \\
\texttt{pre\_conditions.append(Exists([m], eats(Vladimir, m) == eats(Wendy, m)))} \\
\texttt{\bf RepairedPositiveExampleCode:} \\
\texttt{NONE} \\
\texttt{------} \\
\texttt{\bf Scenario:} \\
\texttt{In a repair facility there are exactly six technicians: Stacy, Urma, Wim, Xena, Yolanda, and Zane. Each technician repairs machines of at least one of the following three types—radios, televisions, and VCRs—and no other types.} \\
\texttt{\bf InitialCode:} \\
\texttt{from z3 import *} \\
\texttt{technicians\_sort, (Stacy, Urma, Wim, Xena, Yolanda, Zane) = EnumSort('technicians', ['Stacy', 'Urma', 'Wim', 'Xena', 'Yolanda', 'Zane'])} \\
\texttt{machines\_sort, (radios, televisions, VCRs) = EnumSort('machines', ['radios', 'televisions', 'VCRs'])} \\
\texttt{technicians = [Stacy, Urma, Wim, Xena, Yolanda, Zane]} \\
\texttt{machines = [radios, televisions, VCRs]} \\
\texttt{repairs = Function('repairs', technicians\_sort, machines\_sort, BoolSort())} \\

\texttt{pre\_conditions = []} \\
\texttt{t = Const('t', technicians\_sort)} \\
\texttt{pre\_conditions.append(ForAll([t], Sum([repairs(t, m) for m in machines]) <= 1))} \\
\texttt{\bf ConstraintDescription:} \\
\texttt{Urma repairs radios and VCRs} \\
\texttt{\bf ConstraintCode:} \\
\texttt{pre\_conditions.append(And(repairs(Urma, radios), repairs(Urma, VCRs)))} \\
\texttt{\bf PositiveExampleCode:} \\
\texttt{And(repairs(Urma, radios) == True, repairs(Urma, VCRs) == True)} \\
\texttt{\bf ProblemDiscussion:} \\
\texttt{The scenario describes types of machines that technicians repair at a repair facility, where each technician repairs at least one type of machine. The initial code defines the main data structures and the repairs function which indicates the type of machine repaired by each technician. It also adds the general condition that each technician can repair at most one type of machine, which is an incorret interpretation of the scenario statement that each technician must repair AT LEAST one type of machine. The constraint requires that Urma repairs both VCRs and radios. The constraint code correctly asserts this requirement, and the positive example code also states this correctly. Hence there is no issue in the constraint code and the example code, but the initial code wrongly prevents any technician from repairing two kinds of machines. It should be repaired to assert that each technician must repair at least one kind of machine.} \\
\texttt{\bf RepairedInitialCode:} \\
\texttt{from z3 import *} \\
\texttt{technicians\_sort, (Stacy, Urma, Wim, Xena, Yolanda, Zane) = EnumSort('technicians', ['Stacy', 'Urma', 'Wim', 'Xena', 'Yolanda', 'Zane'])} \\
\texttt{machines\_sort, (radios, televisions, VCRs) = EnumSort('machines', ['radios', 'televisions', 'VCRs'])} \\
\texttt{technicians = [Stacy, Urma, Wim, Xena, Yolanda, Zane]} \\
\texttt{machines = [radios, televisions, VCRs]} \\
\texttt{repairs = Function('repairs', technicians\_sort, machines\_sort, BoolSort())} \\

\texttt{pre\_conditions = []} \\
\texttt{t = Const('t', technicians\_sort)} \\
\texttt{pre\_conditions.append(ForAll([t], Sum([repairs(t, m) for m in machines]) >= 1))} \\
\texttt{\bf RepairedConstraintCode:} \\
\texttt{NONE} \\
\texttt{\bf RepairedPositiveExampleCode:} \\
\texttt{NONE} \\
\texttt{------} \\

\subsection{Dataset correction cases}
\label{app:dataset-corrections}

We found a small number of cases in three of the datasets where the answers have been labelled incorrectly. Our SSV system (with GPT-4 base model) detected these cases in its verification, and we describe the corrections that should be made to the datasets below. 

\subsubsection{AR-LSAT Corrections}

Three cases in the AR-LSAT dataset were verified correctly by our system, but were labelled with the wrong answers in the dataset. These three cases are {\bf ar\_lsat\_201612\_3-G\_2\_6} (correct answer should be D but incorrectly labelled C), {\bf ar\_lsat\_201612\_3-G\_1\_4} (correct answer should be E but incorrectly labelled A) and {\bf ar\_lsat\_201612\_3-G\_2\_8} (correct answer should be B but is incorrectly labelled A). For all three of these cases, we were able to check the reasoning and also that the answers in the original source LSAT Test 
({\href{https://img.cracklsat.net/lsat/pt/pt80.pdf}{https://img.cracklsat.net/lsat/pt/pt80.pdf}}) are consistent with the answers that were generated by our system. Hence we submit that these are errors in the AR-LSAT dataset collection process.

\subsubsection{FOLIO Corrections}

In the FOLIO dataset, we found one case that was correctly verified by our system, but we find is labelled with the wrong answer in the dataset. This is case {\bf FOLIO\_dev\_27}:

\emph{All aliens are extraterrestrial. If someone is from Mars, then they are aliens. No extraterrestrial is human. Everyone from Earth is a human. Marvin cannot be from Earth and from Mars. If Marvin is not from Earth, then Marvin is an extraterrestrial. Based on the above information, is the following statement true, false, or uncertain? Marvin is an alien.}

We submit that the correct answer is C (unknown) but it is labelled B (false) in the dataset. Reasoning:  If Marvin is from Earth, he is not an alien. If Marvin is not from Earth: If he is from Mars, he is an alien, otherwise, we cannot be certain he is an alien. Hence both outcomes are possible. 

We suspect the error in the dataset may stem from an incorrect formalization of the problem in the original FOLIO dataset source:\href{https://github.com/Yale-LILY/FOLIO/blob/main/data/v0.0/folio-validation.txt}{https://github.com/Yale-LILY/FOLIO/blob/main/data/v0.0/folio-validation.txt}. In this source we see that the constraint ``Marvin cannot be from Earth and from Mars" is incorrectly formalized as $\neg FromEarth(marvin) \wedge \neg FromMars(marvin)$  in first order logic, which asserts that Marvin is neither from Earth nor from Mars. 

\subsubsection{ProofWriter corrections}

In the ProofWriter dataset, we found 6 cases that were correctly  verified by our system,  but we find  are labelled with the wrong answer in the dataset. In all 6 cases, the answers in the dataset have been labelled as unknown when they can be proven to be either true or false as we show below. 

{\bf ProofWriter\_RelNeg-OWA-D5-450\_Q22 }  (Correct answer should be B (false), but labelled C (unknown)).

\emph{The bald eagle chases the lion. The bald eagle is not green. The bald eagle is round. The bald eagle likes the lion. The dog is red. The lion does not chase the dog. The lion is round. The lion is not young. The rabbit chases the dog. The rabbit eats the lion. If something chases the dog then it likes the rabbit. If something is red and it chases the lion then the lion likes the bald eagle. If something is big then it chases the rabbit. If something is round and it chases the bald eagle then the bald eagle does not like the dog. If something likes the lion then it is red. If something is red and round then it does not chase the bald eagle. If something is red and young then it chases the bald eagle. If something likes the bald eagle and the bald eagle chases the lion then it likes the lion. If something eats the bald eagle then the bald eagle is red. Based on the above information, is the following statement true, false, or unknown? The bald eagle is young.}

Reasoning:

From Fact 4 and Rule 5:

The bald eagle likes the lion.
Therefore, the bald eagle is red.

From Fact 3:

The bald eagle is round.
Applying Rule 6 to the bald eagle:

The bald eagle is red and round.
Therefore, the bald eagle does not chase itself.
Assuming the bald eagle is young:

The bald eagle is red and young.
Applying Rule 7 to the bald eagle:

The bald eagle is red and young.
Therefore, the bald eagle chases itself.
Contradiction:

From step 3, the bald eagle does not chase itself.

From step 5, the bald eagle chases itself.

This is a contradiction.

Conclusion:
The assumption that the bald eagle is young leads to a contradiction.
Therefore, the bald eagle cannot be young.

\paragraph{ProofWriter\_AttNeg-OWA-D5-471\_Q14 }  (Correct answer should be A (true), but labelled C (unknown)).

\emph{Anne is white. Charlie is cold. Charlie is round. Charlie is young. Gary is kind. Gary is nice. Gary is round. Gary is white. Gary is young. Harry is blue. Harry is cold. Harry is kind. Harry is white. Harry is young. White, kind things are blue. If something is white then it is kind. Nice things are kind. All blue, nice things are young. All blue, white things are nice. If something is round and not nice then it is not cold. Blue, young things are cold. Based on the above information, is the following statement true, false, or unknown? Charlie is kind.}

Reasoning: 

Relevant facts: Charlie is cold.
Charlie is round.
Charlie is young.

Relevant Rules:

If something is round and not nice, then it is not cold. (Rule 6)

Nice things are kind. (Rule 3)

Assuming Charlie is not nice:

Since Charlie is round and not nice, according to Rule 6, Charlie should not be cold.
However, this contradicts the fact that Charlie is cold. Therefore, our assumption that Charlie is not nice must be false.

Conclusion from the contradiction:
Charlie must be nice.

Applying Rule 3:

Since nice things are kind, and Charlie is nice, it follows that Charlie is kind.

\paragraph{ProofWriter\_AttNeg-OWA-D5-112\_Q20}  (Correct answer should be B (false), but labelled C (unknown)).

\emph{Charlie is kind. Charlie is nice. Charlie is quiet. Dave is rough. Dave is white. Erin is nice. Gary is not white. If something is cold then it is not furry. If Charlie is quiet then Charlie is nice. Kind things are white. Nice things are kind. If something is rough then it is kind. Cold, quiet things are rough. All cold things are quiet. If something is white and nice then it is cold. If Erin is cold then Erin is nice. Based on the above information, is the following statement true, false, or unknown? Gary is nice.}

Reasoning:

Gary is not white. (rule 1)

Nice things are kind. (rule 2)

Kind things are white. (rule 3)

If Gary were nice, then by rule 2, he would also be kind.
If Gary is kind, then by rule 3, he must be white.
However, rule 1 tells us that Gary is not white.
This creates a contradiction because Gary cannot be both not white and white at the same time.

Given that Gary is not white, he cannot be kind, and therefore, he cannot be nice. Thus, the statement ``Gary is nice" is false.

\paragraph{ProofWriter\_AttNeg-OWA-D5-850\_Q14}  (Correct answer should be B (false), but labelled C (unknown)).

\emph{Anne is red. Anne is smart. Bob is kind. Bob is not nice. Fiona is furry. Fiona is rough. Gary is not green. Gary is kind. Gary is nice. Gary is rough. If someone is nice then they are red. Smart people are green. If someone is smart and red then they are not kind. All rough, green people are nice. Green people are rough. If someone is red and green then they are rough. If someone is furry and green then they are smart. All rough, furry people are smart. Furry, rough people are smart. Based on the above information, is the following statement true, false, or unknown? Bob is smart.}

Reasoning:

Bob is kind. Bob is not nice.

Rule: Smart people are green. So, if Bob were smart, he would be green.

Rule: Green people are rough. Therefore, if Bob were green (and thus rough), we can use the next rule.

Rule: All rough, green people are nice. If Bob were rough and green, he would be nice, but we know Bob is not nice.

Conclusion: Bob cannot be green because it would contradict the fact that he is not nice.
Since Bob is not green, and smart people are green, Bob cannot be smart.

\paragraph{ProofWriter\_AttNeg-OWA-D5-219\_Q13}  (Correct answer should be A (true), but labelled C (unknown)).

\emph{Charlie is not quiet. Dave is big. Dave is furry. Erin is cold. Erin is not green. Erin is not kind. Fiona is quiet. Big things are young. Young, cold things are big. Quiet things are big. All young things are cold. If something is big and not furry then it is cold. If something is cold then it is not kind. If something is cold and big then it is quiet. If Fiona is cold and Fiona is not quiet then Fiona is kind. If something is quiet and not kind then it is green. Based on the above information, is the following statement true, false, or unknown? Charlie is not big.}

Reasoning:

Charlie is not quiet.

Assume for contradiction that Charlie is big.

Big things are young: Therefore, Charlie is young.

All young things are cold: Therefore, Charlie is cold.

If something is cold, then it is not kind: Therefore, Charlie is not kind.

If something is cold and big, then it is quiet: Therefore, Charlie is quiet.

This contradicts the given fact that Charlie is not quiet. Therefore, Charlie is not big.

\paragraph{ProofWriter\_AttNeg-OWA-D5-94\_Q18}   (Correct answer should be B (false), but labelled C (unknown))

\emph{Bob is smart. Charlie is kind. Charlie is not smart. Fiona is blue. Fiona is rough. Fiona is smart. Gary is kind. All cold, quiet people are smart. If someone is cold then they are smart. If someone is red and kind then they are smart. If someone is quiet then they are blue. If someone is blue then they are quiet. If someone is kind then they are rough. If Gary is kind and Gary is rough then Gary is quiet. All blue, smart people are red. Blue, rough people are red. Based on the above information, is the following statement true, false, or unknown? Charlie is blue.}

Reasoning:

Charlie is kind.

If someone is kind, then they are rough 

Therefore, Charlie is rough.

Assume for contradiction that Charlie is blue.

Blue, rough people are red. Since Charlie is both blue (assumed) and rough, Charlie must be red.

If someone is red and kind, then they are smart. 

Since Charlie is red (from step 4) and kind (from step 1), Charlie must be smart.

However, it's given that Charlie is not smart (from the context).

Hence, we have a contradiction. Therefore, Charlie is not blue.

\subsection{Analysis of Verification Failure Cases}
\label{app:failanalysis}

We conducted a manual analysis over a sample of 30 cases where SSV verification failed. Here is a summary of the failure reasons:
\begin{itemize}
    \item code incorrect, example correct: 16 (53.3\%)
    \item code incorrect, example incorrect: 7 (23.3\%)
	\item code correct, example incorrect: 3 (10\%)
    \item program not well-formed: 4 (13.3\%)
\end{itemize}
	
We see that in most cases the code is incorrect as opposed to examples, which can be expected as examples inference is generally simpler than abstract translation. Below is the detailed analysis of the reasons for the verification failure for specific cases.

\textbf{InitialContext:}
\begin{Verbatim}
A bakery makes exactly three kinds of cookie—oatmeal, peanut butter, and sugar. 
Exactly three batches of each kind of cookie are made each week (Monday through Friday) 
and each batch is made, from start to finish, on a single day.
\end{Verbatim}

\textbf{InitialCode:}
\begin{Verbatim}
from z3 import *

days_sort, (Monday, Tuesday, Wednesday, Thursday, Friday) = EnumSort('days', ['Monday', 'Tuesday', 'Wednesday', 'Thursday', 'Friday'])
cookies_sort, (oatmeal, peanut_butter, sugar) = EnumSort('cookies', ['oatmeal', 'peanut_butter', 'sugar'])
batch_number_sort = IntSort()
make_batch = Function('make_batch', cookies_sort, batch_number_sort, days_sort)

pre_conditions = []
batches = range(1, 3+1)  # Corrected to only three batches
days = [Monday, Tuesday, Wednesday, Thursday, Friday]
cookies = [oatmeal, peanut_butter, sugar]

for cookie in cookies:
    for b in batches:
        d = Const('d', days_sort)  # Declare a constant of type days_sort
        pre_conditions.append(Exists([d], make_batch(cookie, b) == d))
\end{Verbatim}

\textbf{NegExample:}
\begin{Verbatim}
And(make_batch(oatmeal, 1) == Monday, make_batch(oatmeal, 2) == Tuesday, make_batch(oatmeal, 3) == Wednesday, make_batch(oatmeal, 4) == Thursday)
\end{Verbatim}

\textbf{NegExampleDescription:}
\begin{Verbatim}
Four batches of oatmeal cookies are made, which contradicts the constraint that exactly three batches of each kind of cookie are made each week.
\end{Verbatim}

\textbf{Analysis:}
\begin{Verbatim}
Code incorrect, Example correct. It does not implement the intended constraint that only 3 batches of each kind of cookie should be made in a week, which the negative example correctly tests.
\end{Verbatim}

\hrulefill

\textbf{InitialContext:}
\begin{Verbatim}
A loading dock consists of exactly six bays numbered 1 through 6 consecutively from one side of the dock to the other. Each bay is holding a different one of exactly six types of cargo—fuel, grain, livestock, machinery, produce, or textiles.
\end{Verbatim}

\textbf{InitialCode:}
\begin{Verbatim}
from z3 import *

bays_sort = IntSort()
cargo_sort, (fuel, grain, livestock, machinery, produce, textiles) = EnumSort('cargo', ['fuel', 'grain', 'livestock', 'machinery', 'produce', 'textiles'])
cargo = [fuel, grain, livestock, machinery, produce, textiles]
holding = Function('holding', bays_sort, cargo_sort)

pre_conditions = []

pre_conditions.append(Distinct([holding(b) for b in range(1, 7)]))
pre_conditions.append(And([holding(b) != None for b in range(1, 7)]))

\end{Verbatim}

\textbf{Condition:}
\begin{Verbatim}
The bay holding textiles is next to the bay holding produce.
\end{Verbatim}

\textbf{ConditionCode:}
\begin{Verbatim}
b = Int('b')
pre_conditions.append(Or(holding(b) == textiles, holding(b+1) == produce, holding(b-1) == produce))
\end{Verbatim}

\textbf{NegExample:}
\begin{Verbatim}
And(holding(1) == textiles, holding(4) == produce)
\end{Verbatim}

\textbf{NegExampleDescription:}
\begin{Verbatim}
Textiles are in bay 1 and produce is in bay 4.
\end{Verbatim}

\textbf{Analysis:}
\begin{Verbatim}
Code incorrect, Example correct. It does not correctly implement the constraint that textiles should be right next to produce as it is only a disjunction of the three conditions. The negative example correctly checks this condition by allowing textiles in bay 1 and produce in bay 4.
\end{Verbatim}

\hrulefill

\textbf{InitialContext:}
\begin{Verbatim}
On each of exactly seven consecutive days (day 1 though day 7), a pet shop features exactly one of three breeds of kitten—Himalayan, Manx, Siamese—and exactly one of three breeds of puppy—Greyhound, Newfoundland, Rottweiler.
\end{Verbatim}

\textbf{InitialCode:}
\begin{Verbatim}
from z3 import *

days_sort, (day1, day2, day3, day4, day5, day6, day7) = EnumSort('days', ['day1', 'day2', 'day3', 'day4', 'day5', 'day6', 'day7'])
kitten_breeds_sort, (Himalayan, Manx, Siamese) = EnumSort('kitten_breeds', ['Himalayan', 'Manx', 'Siamese'])
puppy_breeds_sort, (Greyhound, Newfoundland, Rottweiler) = EnumSort('puppy_breeds', ['Greyhound', 'Newfoundland', 'Rottweiler'])
days = [day1, day2, day3, day4, day5, day6, day7]
kitten_breeds = [Himalayan, Manx, Siamese]
puppy_breeds = [Greyhound, Newfoundland, Rottweiler]

features_kitten = Function('features_kitten', days_sort, kitten_breeds_sort, BoolSort())
features_puppy = Function('features_puppy', days_sort, puppy_breeds_sort, BoolSort())

pre_conditions = []

# Add constraints to ensure exactly one breed of kitten and one breed of puppy is featured each day
for day in days:
    pre_conditions.append(Sum([If(features_kitten(day, breed), 1, 0) for breed in kitten_breeds]) == 1)
    pre_conditions.append(Sum([If(features_puppy(day, breed), 1, 0) for breed in puppy_breeds]) == 1)
\end{Verbatim}

\textbf{Condition:}
\begin{Verbatim}
No breed is featured on any two consecutive days.
\end{Verbatim}

\textbf{ConditionCode:}
\begin{Verbatim}
d1 = Const('d1', days_sort)
d2 = Const('d2', days_sort)
kb = Const('kb', kitten_breeds_sort)
pb = Const('pb', puppy_breeds_sort)
pre_conditions.append(ForAll([d1, d2, kb], Implies(And(features_kitten(d1, kb), features_kitten(d2, kb)), d1 == d2)))
pre_conditions.append(ForAll([d1, d2, pb], Implies(And(features_puppy(d1, pb), features_puppy(d2, pb)), d1 == d2)))
\end{Verbatim}

\textbf{PosExample:}
\begin{Verbatim}
And(features_kitten(day2, Siamese) == True, features_kitten(day3, Manx) == True)
\end{Verbatim}

\textbf{PosExampleDescription:}
\begin{Verbatim}
Siamese kittens are featured on day 2 and Manx kittens are featured on day 3.
\end{Verbatim}

\textbf{Analysis:}
\begin{Verbatim}
Code incorrect, Example correct. The condition code does not implement consecutiveness constraint but instead requires each breed to only be featured on one day, which is not a requirement and causes the positive example to fail. The positive example correctly presents a valid instantiation of the problem.
\end{Verbatim}

\hrulefill

\textbf{InitialContext:}
\begin{Verbatim}
The organizer of a reading club will select at least five and at most six works from a group of nine works. The group consists of three French novels, three Russian novels, two French plays, and one Russian play.
\end{Verbatim}

\textbf{InitialCode:}
\begin{Verbatim}
from z3 import *

works_sort, (french_novel1, french_novel2, french_novel3, russian_novel1, russian_novel2, russian_novel3, french_play1, french_play2, russian_play) = EnumSort('works', ['french_novel1', 'french_novel2', 'french_novel3', 'russian_novel1', 'russian_novel2', 'russian_novel3', 'french_play1', 'french_play2', 'russian_play'])
works = [french_novel1, french_novel2, french_novel3, russian_novel1, russian_novel2, russian_novel3, french_play1, french_play2, russian_play]
selected = Function('selected', works_sort, BoolSort())

pre_conditions = []
\end{Verbatim}

\textbf{NegExample:}
\begin{Verbatim}
And(selected(french_novel1) == True, selected(french_novel2) == True, selected(french_novel3) == True, selected(russian_novel1) == True, selected(russian_novel2) == True, selected(russian_novel3) == True, selected(french_play1) == True, selected(french_play2) == True, selected(russian_play) == True)
\end{Verbatim}

\textbf{NegExampleDescription:}
\begin{Verbatim}
The organizer selects all three French novels, all three Russian novels, both French plays, and the Russian play, which exceeds the maximum of six works.
\end{Verbatim}

\textbf{Analysis:}
\begin{Verbatim}
Code incorrect, Example correct. The negative example correctly violates the constraint of not exceeding 6 works, but the code does not implement this constraint.
\end{Verbatim}

\hrulefill

\textbf{InitialContext:}
\begin{Verbatim}
A bakery makes exactly three kinds of cookie—oatmeal, peanut butter, and sugar. 
Exactly three batches of each kind of cookie are made each week (Monday through Friday) 
and each batch is made, from start to finish, on a single day.
\end{Verbatim}

\textbf{InitialCode:}
\begin{Verbatim}
from z3 import *

days_sort, (Monday, Tuesday, Wednesday, Thursday, Friday) = EnumSort('days', ['Monday', 'Tuesday', 'Wednesday', 'Thursday', 'Friday'])
cookies_sort, (oatmeal, peanut_butter, sugar) = EnumSort('cookies', ['oatmeal', 'peanut_butter', 'sugar'])
batch_sort = IntSort()
make_batch = Function('make_batch', cookies_sort, batch_sort, days_sort, BoolSort())
pre_conditions = []

# Add constraints to ensure exactly three batches of each kind of cookie are made each week
for cookie in [oatmeal, peanut_butter, sugar]:
    pre_conditions.append(Sum([If(make_batch(cookie, i, d), 1, 0) for i in range(1, 4) for d in [Monday, Tuesday, Wednesday, Thursday, Friday]]) == 3)
\end{Verbatim}

\textbf{NegExample:}
\begin{Verbatim}
And(make_batch(oatmeal, 1, Monday) == True, make_batch(oatmeal, 2, Tuesday) == True, make_batch(oatmeal, 3, Wednesday) == True, make_batch(oatmeal, 4, Thursday) == True)
\end{Verbatim}

\textbf{NegExampleDescription:}
\begin{Verbatim}
Four batches of oatmeal cookies are made, which contradicts the constraint that exactly three batches of each kind of cookie are made each week.
\end{Verbatim}

\textbf{Analysis:}
\begin{Verbatim}
Code incorrect, Example correct. The negative example correctly violates the constraint by enforcing 4 batches oatmeal cookies to be made in the week.
\end{Verbatim}

\hrulefill

\textbf{InitialContext:}
\begin{Verbatim}
An administrator must assign parking spaces to six new employees: Robertson, Souza, Togowa, Vaughn, Xu, and Young. 
Each of the six employees must be assigned one of the following parking spaces: #1, #2, #3, #4, #5, or #6. 
No two employees can be assigned the same parking space.
\end{Verbatim}

\textbf{InitialCode:}
\begin{Verbatim}
from z3 import *

employees_sort, (Robertson, Souza, Togowa, Vaughn, Xu, Young) = EnumSort('employees', ['Robertson', 'Souza', 'Togowa', 'Vaughn', 'Xu', 'Young'])
parking_spaces_sort = IntSort()
parking_space = Function('parking_space', employees_sort, parking_spaces_sort)
pre_conditions = []

pre_conditions.append(Distinct([parking_space(e) for e in [Robertson, Souza, Togowa, Vaughn, Xu, Young]]))
pre_conditions.append(And([parking_space(e) >= 1 for e in [Robertson, Souza, Togowa, Vaughn, Xu, Young]]))
pre_conditions.append(And([parking_space(e) <= 6 for e in [Robertson, Souza, Togowa, Vaughn, Xu, Young]]))

\end{Verbatim}

\textbf{Condition:}
\begin{Verbatim}
If Togowa is assigned a higher-numbered parking space than Souza, then Togowa is assigned parking space #3.
\end{Verbatim}

\textbf{ConditionCode:}
\begin{Verbatim}
additional_constraint = parking_space(Togowa) > parking_space(Souza)
check_property = And(additional_constraint, parking_space(Togowa) == 3)
pre_conditions.append(check_property)
\end{Verbatim}

\textbf{ExampleFail:}
\begin{Verbatim}
Positive example not inferred (NONE description)
\end{Verbatim}

\textbf{Analysis:}
\begin{Verbatim}
Code incorrect, Example incorrect. The code does not implement the conditional requirement but only a conjunction. A positive example could not be inferred from the LLM or valid solver code implementing the positive example.
\end{Verbatim}

\hrulefill

\textbf{InitialContext:}
\begin{Verbatim}
A panel of five scientists will be formed. The panelists will be selected from among three botanists—F, G, and H—three chemists—K, L, and M—and three zoologists—P, Q, and R.
\end{Verbatim}

\textbf{InitialCode:}
\begin{Verbatim}
from z3 import *

scientist_sort = DeclareSort('scientist')
F, G, H, K, L, M, P, Q, R = Consts('F G H K L M P Q R', scientist_sort)
botanists = [F, G, H]
chemists = [K, L, M]
zoologists = [P, Q, R]
selected = Function('selected', scientist_sort, BoolSort())

pre_conditions = []
\end{Verbatim}

\textbf{ExampleFail:}
\begin{Verbatim}
Positive example contains undeclared free variables
\end{Verbatim}

\textbf{Analysis:}
\begin{Verbatim}
Code incorrect, Example incorrect. Code is correct but does not implement a constraint that exactly 5 scientists must be selected. A valid positive example in solver code could not be generated.
\end{Verbatim}

\hrulefill

\textbf{InitialContext:}
\begin{Verbatim}
At a concert, exactly eight compositions—F, H, L, O, P, R, S, and T—are to be performed exactly once each, consecutively and one composition at a time.
\end{Verbatim}

\textbf{InitialCode:}
\begin{Verbatim}
from z3 import *

compositions_sort, (F, H, L, O, P, R, S, T) = EnumSort('compositions', ['F', 'H', 'L', 'O', 'P', 'R', 'S', 'T'])
compositions = [F, H, L, O, P, R, S, T]
positions_sort = IntSort()
position = Function('position', compositions_sort, positions_sort)
pre_conditions = []

pre_conditions.append(Distinct([position(c) for c in compositions]))
pre_conditions.append(And([position(c) >= 1 for c in compositions]))
pre_conditions.append(And([position(c) <= 8 for c in compositions]))

\end{Verbatim}

\textbf{Condition:}
\begin{Verbatim}
T is performed either immediately before F or immediately after R.
\end{Verbatim}

\textbf{ConditionCode:}
\begin{Verbatim}
pre_conditions.append(Or(position(T) + 1 == position(F), position(T) == position(R) + 1))
\end{Verbatim}

\textbf{NegExample:}
\begin{Verbatim}
position(T) == position(F) - 3
\end{Verbatim}

\textbf{NegExampleDescription:}
\begin{Verbatim}
T is performed with two compositions between it and F.
\end{Verbatim}

\textbf{Analysis:}
\begin{Verbatim}
Code correct, Example incorrect. The negative example only requires that T is 3 positions before F, which is permitted by the constraint as long as T is immediately after R (since it was a disjunction).
\end{Verbatim}

\hrulefill

\textbf{InitialContext:}
\begin{Verbatim}
A government needs to assign new ambassadors to Venezuela, Yemen, and Zambia. The candidates for these ambassadorships are Jaramillo, Kayne, Landon, Novetzke, and Ong. One ambassador will be assigned to each country, and no ambassador will be assigned to more than one country.
\end{Verbatim}

\textbf{InitialCode:}
\begin{Verbatim}
from z3 import *

candidates_sort, (Jaramillo, Kayne, Landon, Novetzke, Ong) = EnumSort('candidates', ['Jaramillo', 'Kayne', 'Landon', 'Novetzke', 'Ong'])
countries_sort, (Venezuela, Yemen, Zambia) = EnumSort('countries', ['Venezuela', 'Yemen', 'Zambia'])
candidates = [Jaramillo, Kayne, Landon, Novetzke, Ong]
countries = [Venezuela, Yemen, Zambia]
assigned_to = Function('assigned_to', candidates_sort, countries_sort, BoolSort())

pre_conditions = []

pre_conditions.append(Distinct([assigned_to(c, country) for c in candidates for country in countries]))
pre_conditions.append(And([Sum([If(assigned_to(c, country), 1, 0) for country in countries]) == 1 for c in candidates]))
pre_conditions.append(And([Sum([If(assigned_to(c, country), 1, 0) for c in candidates]) == 1 for country in countries]))

\end{Verbatim}

\textbf{PosExample:}
\begin{Verbatim}
And(assigned_to(Jaramillo, Venezuela) == True, assigned_to(Kayne, Yemen) == True, assigned_to(Landon, Zambia) == True, assigned_to(Novetzke, Venezuela) == False, assigned_to(Novetzke, Yemen) == False, assigned_to(Novetzke, Zambia) == False, assigned_to(Ong, Venezuela) == False, assigned_to(Ong, Yemen) == False, assigned_to(Ong, Zambia) == False)
\end{Verbatim}

\textbf{PosExampleDescription:}
\begin{Verbatim}
Jaramillo is assigned to Venezuela, Kayne to Yemen, and Landon to Zambia, while Novetzke and Ong are not assigned to any country.
\end{Verbatim}

\textbf{Analysis:}
\begin{Verbatim}
Code incorrect, Example correct. The positive example correctly instantiates the problem context, but the code incorrectly enforces the very strong distinctness constraint on the assigned_to() function which has boolean return type and should not have dictinct outputs over all possible assignments.
\end{Verbatim}

\hrulefill

\textbf{InitialContext:}
\begin{Verbatim}
Of the eight students—George, Helen, Irving, Kyle, Lenore, Nina, Olivia, and Robert—in a seminar, exactly six will give individual oral reports during three consecutive days—Monday, Tuesday, and Wednesday. Exactly two reports will be given each day—one in the morning and one in the afternoon.
\end{Verbatim}

\textbf{InitialCode:}
\begin{Verbatim}
from z3 import *

students_sort, (George, Helen, Irving, Kyle, Lenore, Nina, Olivia, Robert) = EnumSort('students', ['George', 'Helen', 'Irving', 'Kyle', 'Lenore', 'Nina', 'Olivia', 'Robert'])
days_sort, (Monday, Tuesday, Wednesday) = EnumSort('days', ['Monday', 'Tuesday', 'Wednesday'])
time_sort, (morning, afternoon) = EnumSort('time', ['morning', 'afternoon'])
students = [George, Helen, Irving, Kyle, Lenore, Nina, Olivia, Robert]
days = [Monday, Tuesday, Wednesday]
times = [morning, afternoon]
report = Function('report', students_sort, days_sort, time_sort, BoolSort())

pre_conditions = []
\end{Verbatim}

\textbf{NegExample:}
\begin{Verbatim}
And(report(George, Monday, morning) == True, report(George, Tuesday, morning) == True)
\end{Verbatim}

\textbf{NegExampleDescription:}
\begin{Verbatim}
George gives two reports, one on Monday morning and another on Tuesday morning, which contradicts the constraint that each student gives only one report.
\end{Verbatim}

\textbf{Analysis:}
\begin{Verbatim}
Code incorrect, Example correct. It does not enforce that a student cannot give multiple reports, so it will not be possible for exactly 6 reports to be given by exactly 6 students as intended. The example correctly violates the constraint by enforcing George to give two reports in different slots.
\end{Verbatim}

\hrulefill

\textbf{InitialContext:}
\begin{Verbatim}
A bakery makes exactly three kinds of cookie—oatmeal, peanut butter, and sugar. 
Exactly three batches of each kind of cookie are made each week (Monday through Friday) 
and each batch is made, from start to finish, on a single day.
\end{Verbatim}

\textbf{InitialCode:}
\begin{Verbatim}
from z3 import *

days_sort, (Monday, Tuesday, Wednesday, Thursday, Friday) = EnumSort('days', ['Monday', 'Tuesday', 'Wednesday', 'Thursday', 'Friday'])
cookie_sort, (oatmeal, peanut_butter, sugar) = EnumSort('cookie', ['oatmeal', 'peanut_butter', 'sugar'])
batch_sort = IntSort()
made_on = Function('made_on', cookie_sort, batch_sort, days_sort)

pre_conditions = []

for cookie in [oatmeal, peanut_butter, sugar]:
    pre_conditions.append(And([Exists([Const('d' + str(i), days_sort)], made_on(cookie, i) == Const('d' + str(i), days_sort)) for i in range(1, 4)]))
    for day in [Monday, Tuesday, Wednesday, Thursday, Friday]:
        pre_conditions.append(Sum([If(made_on(cookie, i) == day, 1, 0) for i in range(1, 4)]) <= 1)

\end{Verbatim}

\textbf{Condition:}
\begin{Verbatim}
If no batch of cookies is made on Wednesday, then exactly three batches of cookies are made on Tuesday.
\end{Verbatim}

\textbf{ConditionCode:}
\begin{Verbatim}
day = Const('day', days_sort)
check_property = Sum([If(made_on(cookie, i) == Tuesday, 1, 0) for cookie in [oatmeal, peanut_butter, sugar] for i in range(1, 4)]) == 3
pre_conditions.append(check_property)
\end{Verbatim}

\textbf{NegExample:}
\begin{Verbatim}
And(made_on(oatmeal, 1) != Wednesday, made_on(peanut_butter, 1) != Wednesday, made_on(sugar, 1) != Wednesday,
made_on(oatmeal, 1) == Tuesday, made_on(peanut_butter, 2) == Tuesday, made_on(sugar, 3) != Tuesday)
\end{Verbatim}

\textbf{NegExampleDescription:}
\begin{Verbatim}
No cookies are made on Wednesday, but only two batches of cookies are made on Tuesday.
\end{Verbatim}

\textbf{Analysis:}
\begin{Verbatim}
Code incorrect, Example incorrect. The code directly enforces that exactly three batches are made on Tuesday, regardless of whether any batches are made on Wednesday so it lacks the conditional aspect "if no batch is made on Wednesday" of the intended constraint. The negative example prevents only batch 3 of sugar cookies on tuesday but not other batches, so it is still possible to have three batches on tuesday.
\end{Verbatim}

\hrulefill

\textbf{InitialContext:}
\begin{Verbatim}
An administrator must assign parking spaces to six new employees: Robertson, Souza, Togowa, Vaughn, Xu, and Young. Each of the six employees must be assigned one of the following parking spaces: #1, #2, #3, #4, #5, or #6. No two employees can be assigned the same parking space.
\end{Verbatim}

\textbf{InitialCode:}
\begin{Verbatim}
from z3 import *

employees_sort, (Robertson, Souza, Togowa, Vaughn, Xu, Young) = EnumSort('employees', ['Robertson', 'Souza', 'Togowa', 'Vaughn', 'Xu', 'Young'])
parking_spaces_sort = IntSort()
parking_space = Function('parking_space', employees_sort, parking_spaces_sort)
pre_conditions = []

pre_conditions.append(Distinct([parking_space(e) for e in [Robertson, Souza, Togowa, Vaughn, Xu, Young]]))
pre_conditions.append(And([parking_space(e) >= 1 for e in [Robertson, Souza, Togowa, Vaughn, Xu, Young]]))
pre_conditions.append(And([parking_space(e) <= 6 for e in [Robertson, Souza, Togowa, Vaughn, Xu, Young]]))

\end{Verbatim}

\textbf{Condition:}
\begin{Verbatim}
If Young is assigned a higher-numbered parking space than Souza, then Young is assigned parking space #2.
\end{Verbatim}

\textbf{ConditionCode:}
\begin{Verbatim}
check_property = And(parking_space(Young) > parking_space(Souza), parking_space(Young) == 2)
pre_conditions.append(check_property)
\end{Verbatim}

\textbf{ExampleFail:}
\begin{Verbatim}
Positive example not inferred (NONE description)
\end{Verbatim}

\textbf{Analysis:}
\begin{Verbatim}
Code incorrect, Example incorrect. The condition implemented just enforces that Young has number 2 and is than Souza, but does not allow for a lower number than Souza as intended. The positive example is not inferred.
\end{Verbatim}

\hrulefill

\textbf{InitialContext:}
\begin{Verbatim}
A bakery makes exactly three kinds of cookie—oatmeal, peanut butter, and sugar. Exactly three batches of each kind of cookie are made each week (Monday through Friday) and each batch is made, from start to finish, on a single day.
\end{Verbatim}

\textbf{InitialCode:}
\begin{Verbatim}
from z3 import *

days_sort, (Monday, Tuesday, Wednesday, Thursday, Friday) = EnumSort('days', ['Monday', 'Tuesday', 'Wednesday', 'Thursday', 'Friday'])
cookies_sort, (oatmeal, peanut_butter, sugar) = EnumSort('cookies', ['oatmeal', 'peanut_butter', 'sugar'])
batch_sort = IntSort()
schedule = Function('schedule', cookies_sort, batch_sort, days_sort)

pre_conditions = []

\end{Verbatim}

\textbf{NegExample:}
\begin{Verbatim}
And(schedule(oatmeal, 1) == Monday, schedule(oatmeal, 2) == Monday, schedule(oatmeal, 3) == Monday, schedule(oatmeal, 4) == Monday)
\end{Verbatim}

\textbf{NegExampleDescription:}
\begin{Verbatim}
Four batches of oatmeal cookies are made on Monday, which contradicts the constraint that exactly three batches of each kind of cookie are made each week.
\end{Verbatim}

\textbf{Analysis:}
\begin{Verbatim}
Code incorrect, Example correct. The code does not implement any constraints that exactly 3 batches of each kind of cookie should be made in a week.
\end{Verbatim}

\hrulefill

\textbf{InitialContext:}
\begin{Verbatim}
A government needs to assign new ambassadors to Venezuela, Yemen, and Zambia. The candidates for these ambassadorships are Jaramillo, Kayne, Landon, Novetzke, and Ong. One ambassador will be assigned to each country, and no ambassador will be assigned to more than one country.
\end{Verbatim}

\textbf{InitialCode:}
\begin{Verbatim}
from z3 import *

ambassadors_sort, (Jaramillo, Kayne, Landon, Novetzke, Ong) = EnumSort('ambassadors', ['Jaramillo', 'Kayne', 'Landon', 'Novetzke', 'Ong'])
countries_sort, (Venezuela, Yemen, Zambia) = EnumSort('countries', ['Venezuela', 'Yemen', 'Zambia'])
ambassadors = [Jaramillo, Kayne, Landon, Novetzke, Ong]
countries = [Venezuela, Yemen, Zambia]
assigned_to = Function('assigned_to', ambassadors_sort, countries_sort, BoolSort())

pre_conditions = []
a = Const('a', ambassadors_sort)
c1 = Const('c1', countries_sort)
c2 = Const('c2', countries_sort)
pre_conditions.append(ForAll([a], Sum([If(assigned_to(a, c), 1, 0) for c in countries]) <= 1))

\end{Verbatim}

\textbf{Condition:}
\begin{Verbatim}
Either Kayne or Novetzke, but not both, is assigned to one of the ambassadorships.
\end{Verbatim}

\textbf{ConditionCode:}
\begin{Verbatim}
pre_conditions.append(Xor(Or(assigned_to(Kayne, Venezuela), assigned_to(Kayne, Yemen), assigned_to(Kayne, Zambia)), Or(assigned_to(Novetzke, Venezuela), assigned_to(Novetzke, Yemen), assigned_to(Novetzke, Zambia))))
\end{Verbatim}

\textbf{ExampleFail:}
\begin{Verbatim}
Positive example contains undeclared free variables
\end{Verbatim}

\textbf{Analysis:}
\begin{Verbatim}
Code correct, Example incorrect. The example is not well formed with undeclared free variables, while the code correctly implements the constraint using exclusive OR.
\end{Verbatim}

\hrulefill

\textbf{InitialContext:}
\begin{Verbatim}
A bakery makes exactly three kinds of cookie—oatmeal, peanut butter, and sugar. Exactly three batches of each kind of cookie are made each week (Monday through Friday) and each batch is made, from start to finish, on a single day.
\end{Verbatim}

\textbf{InitialCode:}
\begin{Verbatim}
from z3 import *

days_sort, (Monday, Tuesday, Wednesday, Thursday, Friday) = EnumSort('days', ['Monday', 'Tuesday', 'Wednesday', 'Thursday', 'Friday'])
cookie_sort, (oatmeal, peanut_butter, sugar) = EnumSort('cookie', ['oatmeal', 'peanut_butter', 'sugar'])
batches_sort = IntSort()
make_batch = Function('make_batch', cookie_sort, batches_sort, days_sort, BoolSort())

pre_conditions = []
days = [Monday, Tuesday, Wednesday, Thursday, Friday]
cookies = [oatmeal, peanut_butter, sugar]

# Constraint: Exactly three batches of each kind of cookie are made each week
for cookie in cookies:
    pre_conditions.append(Sum([If(make_batch(cookie, i, d), 1, 0) for i in range(1, 4) for d in days]) == 3)

\end{Verbatim}

\textbf{NegExample:}
\begin{Verbatim}
And(make_batch(oatmeal, 1, Monday), make_batch(oatmeal, 2, Tuesday), make_batch(oatmeal, 3, Wednesday), make_batch(oatmeal, 4, Thursday))
\end{Verbatim}

\textbf{NegExampleDescription:}
\begin{Verbatim}
Four batches of oatmeal cookies are made on Monday, Tuesday, Wednesday, and Thursday (which contradicts the constraint that exactly three batches of each kind of cookie are made each week).
\end{Verbatim}

\textbf{Analysis:}
\begin{Verbatim}
Code incorrect, Example correct. The code only enforces the constraint for batch numbers 1,2 and 3, but does not restrict any other batch numbers from being created on a day. The negative example uses batch number 4 which is not prevented by the code.
\end{Verbatim}

\hrulefill

\textbf{InitialContext:}
\begin{Verbatim}
A bakery makes exactly three kinds of cookie—oatmeal, peanut butter, and sugar. Exactly three batches of each kind of cookie are made each week (Monday through Friday) and each batch is made, from start to finish, on a single day.
\end{Verbatim}

\textbf{InitialCode:}
\begin{Verbatim}
from z3 import *

days_sort, (Monday, Tuesday, Wednesday, Thursday, Friday) = EnumSort('days', ['Monday', 'Tuesday', 'Wednesday', 'Thursday', 'Friday'])
cookies_sort, (oatmeal, peanut_butter, sugar) = EnumSort('cookies', ['oatmeal', 'peanut_butter', 'sugar'])
batch_sort = IntSort()
make_batch = Function('make_batch', cookies_sort, batch_sort, days_sort)

pre_conditions = []

\end{Verbatim}

\textbf{NegExample:}
\begin{Verbatim}
And(make_batch(oatmeal, 1) == Monday, make_batch(oatmeal, 2) == Monday, make_batch(oatmeal, 3) == Monday, make_batch(oatmeal, 4) == Monday)
\end{Verbatim}

\textbf{NegExampleDescription:}
\begin{Verbatim}
Four batches of oatmeal cookies are made on Monday, which contradicts the constraint that exactly three batches of each kind of cookie are made each week.
\end{Verbatim}

\textbf{Analysis:}
\begin{Verbatim}
Code incorrect, Example correct. The code does not implement any constraint to require exactly 3 batches of each kind of cookie in a week.
\end{Verbatim}

\hrulefill

\textbf{InitialContext:}
\begin{Verbatim}
Of the eight students—George, Helen, Irving, Kyle, Lenore, Nina, Olivia, and Robert—in a seminar, 
exactly six will give individual oral reports during three consecutive days—Monday, Tuesday, and Wednesday. 
Exactly two reports will be given each day—one in the morning and one in the afternoon.
\end{Verbatim}

\textbf{InitialCode:}
\begin{Verbatim}
from z3 import *

students_sort, (George, Helen, Irving, Kyle, Lenore, Nina, Olivia, Robert) = EnumSort('students', 
                           ['George', 'Helen', 'Irving', 'Kyle', 'Lenore', 'Nina', 'Olivia', 'Robert'])
days_sort, (Monday, Tuesday, Wednesday) = EnumSort('days', ['Monday', 'Tuesday', 'Wednesday'])
time_sort, (morning, afternoon) = EnumSort('time', ['morning', 'afternoon'])
students = [George, Helen, Irving, Kyle, Lenore, Nina, Olivia, Robert]
days = [Monday, Tuesday, Wednesday]
times = [morning, afternoon]
report = Function('report', students_sort, days_sort, time_sort, BoolSort())

pre_conditions = []

s = Const('s', students_sort)
d = Const('d', days_sort)
pre_conditions.append(ForAll([s, d], AtMost(report(s, d, morning), report(s, d, afternoon), 1)))

for day in days:
    pre_conditions.append(Sum([If(report(s, day, morning), 1, 0) for s in students]) == 1)
    pre_conditions.append(Sum([If(report(s, day, afternoon), 1, 0) for s in students]) == 1)

pre_conditions.append(Sum([If(Or(report(s, d, morning), report(s, d, afternoon)), 1, 0) for s in students for d in days]) == 6)
\end{Verbatim}

\textbf{Condition:}
\begin{Verbatim}
Helen, Kyle, and Lenore, not necessarily in that order, give the three morning reports.
\end{Verbatim}

\textbf{ConditionCode:}
\begin{Verbatim}
helen_day = Const('helen_day', days_sort)
kyle_day = Const('kyle_day', days_sort)
lenore_day = Const('lenore_day', days_sort)
pre_conditions.append(And(
    report(Helen, helen_day, morning),
    report(Kyle, kyle_day, morning),
    report(Lenore, lenore_day, morning),
    Distinct([helen_day, kyle_day, lenore_day])
))
\end{Verbatim}

\textbf{NegExample:}
\begin{Verbatim}
And(report(Helen, Monday, morning) == True, report(Kyle, Tuesday, afternoon) == True)
\end{Verbatim}

\textbf{NegExampleDescription:}
\begin{Verbatim}
Helen gives a report on Monday morning, but Kyle gives his report in the afternoon.
\end{Verbatim}

\textbf{Analysis:}
\begin{Verbatim}
Code incorrect, Example correct. The code lacks a constraint that ensures each student gives at most one report during the entire period, 
so it allows Kyle to give a morning and an afternoon report. Since this should not be allowed, the negative example correctly violates the requirement 
that Kyle gives a morning report by requiring him to give an afternoon report on Tuesday.
\end{Verbatim}

\hrulefill

\textbf{InitialContext:}
\begin{Verbatim}
A loading dock consists of exactly six bays numbered 1 through 6 consecutively from one side of the dock to the other. 
Each bay is holding a different one of exactly six types of cargo—fuel, grain, livestock, machinery, produce, or textiles.
\end{Verbatim}

\textbf{InitialCode:}
\begin{Verbatim}
from z3 import *

bays_sort = IntSort()
cargo_sort, (fuel, grain, livestock, machinery, produce, textiles) = EnumSort('cargo', 
                      ['fuel', 'grain', 'livestock', 'machinery', 'produce', 'textiles'])
cargo = [fuel, grain, livestock, machinery, produce, textiles]
holding = Function('holding', bays_sort, cargo_sort)

pre_conditions = []

pre_conditions.append(Distinct([holding(b) for b in range(1, 7)]))
pre_conditions.append(And([holding(b) != None for b in range(1, 7)]))
\end{Verbatim}

\textbf{Condition:}
\begin{Verbatim}
The bay holding livestock has a higher number than the bay holding textiles.
\end{Verbatim}

\textbf{ConditionCode:}
\begin{Verbatim}
b1 = Const('b1', bays_sort)
b2 = Const('b2', bays_sort)
pre_conditions.append(Exists([b1], And(holding(b1) == livestock, b1 > 0, ForAll([b2], 
                    Implies(holding(b2) == textiles, b1 > b2)))))
\end{Verbatim}

\textbf{NegExample:}
\begin{Verbatim}
And(holding(5) == textiles, holding(1) == livestock)
\end{Verbatim}

\textbf{NegExampleDescription:}
\begin{Verbatim}
Textiles are in bay 5 and livestock is in bay 1.
\end{Verbatim}

\textbf{Analysis:}
\begin{Verbatim}
Code incorrect, Example correct. The variables b1 and b2 in the constraint code representing bay numbers are not properly constrained 
to be within the valid range of bay numbers (1 through 6). This allows the solver to consider bay numbers outside this range, 
which leads to the negative example not violating the constraint as expected.
\end{Verbatim}

\hrulefill

\textbf{InitialContext:}
\begin{Verbatim}
A bakery makes exactly three kinds of cookie—oatmeal, peanut butter, and sugar. 
Exactly three batches of each kind of cookie are made each week (Monday through Friday) and each batch is made, from start to finish, on a single day.
\end{Verbatim}

\textbf{InitialCode:}
\begin{Verbatim}
from z3 import *

days_sort, (Monday, Tuesday, Wednesday, Thursday, Friday) = EnumSort('days', ['Monday', 'Tuesday', 'Wednesday', 'Thursday', 'Friday'])
cookie_sort, (oatmeal, peanut_butter, sugar) = EnumSort('cookie', ['oatmeal', 'peanut_butter', 'sugar'])
batch_sort = IntSort()
made_on = Function('made_on', cookie_sort, batch_sort, days_sort)

pre_conditions = []

for cookie in [oatmeal, peanut_butter, sugar]:
    pre_conditions.append(Distinct([made_on(cookie, i) for i in range(1, 4)]))
\end{Verbatim}

\textbf{Condition:}
\begin{Verbatim}
Exactly three batches of each kind of cookie are made each week (Monday through Friday) and each batch is made, from start to finish, on a single day.
\end{Verbatim}

\textbf{ConditionCode:}
\begin{Verbatim}
d = Const('d', days_sort)
c = Const('c', cookie_sort)
pre_conditions.append(ForAll([c], Sum([made_on(c, i) == d for i in range(1, 4) for d in [Monday, Tuesday, Wednesday, Thursday, Friday]]) == 3))
\end{Verbatim}

\textbf{ExampleFail:}
\begin{Verbatim}
Positive example contains undeclared free variables
\end{Verbatim}

\textbf{Analysis:}
\begin{Verbatim}
Code incorrect, Example incorrect. The code is incorrect as it only enforces the condition on batch numbers 1..3, but allows any int values for batch numbers, 
so more than 3 batches can potentially be made. The example is incorrect as it contains undeclared free variables.
\end{Verbatim}

\hrulefill

\textbf{InitialContext:}
\begin{Verbatim}
On each of exactly seven consecutive days (day 1 though day 7), a pet shop features exactly one of three breeds of kitten—Himalayan, Manx, Siamese—and exactly one of three breeds of puppy—Greyhound, Newfoundland, Rottweiler.
\end{Verbatim}

\textbf{InitialCode:}
\begin{Verbatim}
from z3 import *

days_sort, (day1, day2, day3, day4, day5, day6, day7) = EnumSort('days', ['day1', 'day2', 'day3', 'day4', 'day5', 'day6', 'day7'])
kitten_breeds_sort, (Himalayan, Manx, Siamese) = EnumSort('kitten_breeds', ['Himalayan', 'Manx', 'Siamese'])
puppy_breeds_sort, (Greyhound, Newfoundland, Rottweiler) = EnumSort('puppy_breeds', ['Greyhound', 'Newfoundland', 'Rottweiler'])
days = [day1, day2, day3, day4, day5, day6, day7]
kitten_breeds = [Himalayan, Manx, Siamese]
puppy_breeds = [Greyhound, Newfoundland, Rottweiler]

featured_kitten = Function('featured_kitten', days_sort, kitten_breeds_sort)
featured_puppy = Function('featured_puppy', days_sort, puppy_breeds_sort)

pre_conditions = []

# Add constraints to ensure exactly one breed of kitten and one breed of puppy is featured each day
d = Const('d', days_sort)
pre_conditions.append(ForAll([d], Sum([If(featured_kitten(d) == breed, 1, 0) for breed in kitten_breeds]) == 1))
pre_conditions.append(ForAll([d], Sum([If(featured_puppy(d) == breed, 1, 0) for breed in puppy_breeds]) == 1))
\end{Verbatim}

\textbf{Condition:}
\begin{Verbatim}
If Himalayans are not featured on day 7, then day 1 and day 3 CANNOT feature both the same breed of kitten and the same breed of puppy.
\end{Verbatim}

\textbf{ConditionCode:}
\begin{Verbatim}
check_property = And(featured_kitten(day1) == featured_kitten(day3), featured_puppy(day1) == featured_puppy(day3), featured_kitten(day7) != Himalayan)
pre_conditions.append(check_property)
\end{Verbatim}

\textbf{PosExample:}
\begin{Verbatim}
And(featured_kitten(day7) != Himalayan, featured_kitten(day1) == Siamese, featured_puppy(day1) == Greyhound, featured_kitten(day3) == Manx, featured_puppy(day3) == Newfoundland)
\end{Verbatim}

\textbf{PosExampleDescription:}
\begin{Verbatim}
Himalayans are not featured on day 7, and day 1 features a Siamese kitten and a Greyhound puppy while day 3 features a Manx kitten and a Newfoundland puppy.
\end{Verbatim}

\textbf{Analysis:}
\begin{Verbatim}
Code incorrect, Example correct. The code implements the condition incorrectly by just enforcing a conjunction of constraints rather than the conditional requirement.
\end{Verbatim}

\hrulefill

\textbf{InitialContext:}
\begin{Verbatim}
A bakery makes exactly three kinds of cookie—oatmeal, peanut butter, and sugar. Exactly three batches of each kind of cookie are made each week (Monday through Friday) and each batch is made, from start to finish, on a single day.
\end{Verbatim}

\textbf{InitialCode:}
\begin{Verbatim}
from z3 import *

days_sort, (Monday, Tuesday, Wednesday, Thursday, Friday) = EnumSort('days', ['Monday', 'Tuesday', 'Wednesday', 'Thursday', 'Friday'])
cookie_sort, (oatmeal, peanut_butter, sugar) = EnumSort('cookie', ['oatmeal', 'peanut_butter', 'sugar'])
batch_sort = IntSort()
made_on = Function('made_on', cookie_sort, batch_sort, days_sort)

pre_conditions = []

# Add constraints to ensure exactly three batches of each kind of cookie are made each week
for cookie in [oatmeal, peanut_butter, sugar]:
    i = Const('i', batch_sort)
    j = Const('j', batch_sort)
    pre_conditions.append(Sum([If(made_on(cookie, k) == day, 1, 0) for k in range(1, 4) for day in [Monday, Tuesday, Wednesday, Thursday, Friday]]) == 3)
    pre_conditions.append(And([Not(Exists([i, j], And(i != j, made_on(cookie, i) == made_on(cookie, j)))) for day in [Monday, Tuesday, Wednesday, Thursday, Friday]]))
\end{Verbatim}

\textbf{PosExample:}
\begin{Verbatim}
And(made_on(oatmeal, 1) == Monday, made_on(oatmeal, 2) == Wednesday, made_on(oatmeal, 3) == Friday,
made_on(peanut_butter, 1) == Tuesday, made_on(peanut_butter, 2) == Thursday, made_on(peanut_butter, 3) == Friday,
made_on(sugar, 1) == Monday, made_on(sugar, 2) == Tuesday, made_on(sugar, 3) == Thursday)
\end{Verbatim}

\textbf{PosExampleDescription:}
\begin{Verbatim}
Three batches of oatmeal cookies are made on Monday, Wednesday, and Friday. Three batches of peanut butter cookies are made on Tuesday, Thursday, and Friday. Three batches of sugar cookies are made on Monday, Tuesday, and Thursday.
\end{Verbatim}

\textbf{Analysis:}
\begin{Verbatim}
Code incorrect, Example correct. The code incorrectly enforces the constraint that a cookie cannot be made on more than one day (which is not intended).
\end{Verbatim}

\hrulefill

\textbf{InitialContext:}
\begin{Verbatim}
A bakery makes exactly three kinds of cookie—oatmeal, peanut butter, and sugar. Exactly three batches of each kind of cookie are made each week (Monday through Friday) and each batch is made, from start to finish, on a single day.
\end{Verbatim}

\textbf{InitialCode:}
\begin{Verbatim}
from z3 import *

days_sort, (Monday, Tuesday, Wednesday, Thursday, Friday) = EnumSort('days', ['Monday', 'Tuesday', 'Wednesday', 'Thursday', 'Friday'])
cookie_sort, (oatmeal, peanut_butter, sugar) = EnumSort('cookie', ['oatmeal', 'peanut_butter', 'sugar'])
batch_sort = IntSort()
made_on = Function('made_on', cookie_sort, batch_sort, days_sort)

pre_conditions = []
\end{Verbatim}

\textbf{NegExample:}
\begin{Verbatim}
And(made_on(oatmeal, 1) == Monday, made_on(oatmeal, 2) == Tuesday, made_on(oatmeal, 3) == Wednesday, made_on(oatmeal, 4) == Thursday)
\end{Verbatim}

\textbf{NegExampleDescription:}
\begin{Verbatim}
Four batches of oatmeal cookies are made, which contradicts the constraint that exactly three batches of each kind of cookie are made each week.
\end{Verbatim}

\textbf{Analysis:}
\begin{Verbatim}
Code incorrect, Example correct. The code does not implement any constraint that enforces that exactly 3 batches of each type of cookie should be made in the week.
\end{Verbatim}

\hrulefill

\textbf{InitialContext:}
\begin{Verbatim}
A bakery makes exactly three kinds of cookie—oatmeal, peanut butter, and sugar. Exactly three batches of 
each kind of cookie are made each week (Monday through Friday) and each batch is made, from start to finish, on a single day.
\end{Verbatim}

\textbf{InitialCode:}
\begin{Verbatim}
from z3 import *

days_sort, (Monday, Tuesday, Wednesday, Thursday, Friday) = EnumSort('days', ['Monday', 'Tuesday', 'Wednesday', 'Thursday', 'Friday'])
cookies_sort, (oatmeal, peanut_butter, sugar) = EnumSort('cookies', ['oatmeal', 'peanut_butter', 'sugar'])
batch_sort = IntSort()
schedule = Function('schedule', cookies_sort, batch_sort, days_sort)

pre_conditions = []

# Constraint: Exactly three batches of each kind of cookie are made each week
for cookie in [oatmeal, peanut_butter, sugar]:
    pre_conditions.append(Sum([If(schedule(cookie, i) == day, 1, 0) for i in range(1, 4) for day in [Monday, Tuesday, Wednesday, Thursday, Friday]]) == 3)
\end{Verbatim}

\textbf{NegExample:}
\begin{Verbatim}
And(schedule(oatmeal, 1) == Monday, schedule(oatmeal, 2) == Tuesday, schedule(oatmeal, 3) == Wednesday, schedule(oatmeal, 4) == Thursday)
\end{Verbatim}

\textbf{NegExampleDescription:}
\begin{Verbatim}
Four batches of oatmeal cookies are made on Monday, Tuesday, Wednesday, and Thursday (which contradicts the constraint that exactly three batches of each kind of cookie are made each week).
\end{Verbatim}

\textbf{Analysis:}
\begin{Verbatim}
Code incorrect, Example correct. The code only implements the constraint for batch numbers 1, 2, and 3, but not for any other batch numbers.
\end{Verbatim}

\hrulefill

\textbf{InitialContext:}
\begin{Verbatim}
On each of exactly seven consecutive days (day 1 though day 7), a pet shop features exactly one of three breeds of kitten—Himalayan, Manx, Siamese—and exactly one of three breeds of puppy—Greyhound, Newfoundland, Rottweiler.
\end{Verbatim}

\textbf{InitialCode:}
\begin{Verbatim}
from z3 import *

days_sort = IntSort()
kitten_breeds_sort, (Himalayan, Manx, Siamese) = EnumSort('kitten_breeds', ['Himalayan', 'Manx', 'Siamese'])
puppy_breeds_sort, (Greyhound, Newfoundland, Rottweiler) = EnumSort('puppy_breeds', ['Greyhound', 'Newfoundland', 'Rottweiler'])

featured_kitten = Function('featured_kitten', days_sort, kitten_breeds_sort)
featured_puppy = Function('featured_puppy', days_sort, puppy_breeds_sort)

pre_conditions = []

pre_conditions.append(And([featured_kitten(i) != featured_kitten(i + 1) for i in range(1, 7)]))
pre_conditions.append(And([featured_puppy(i) != featured_puppy(i + 1) for i in range(1, 7)]))

\end{Verbatim}

\textbf{Condition:}
\begin{Verbatim}
No breed is featured on any two consecutive days.
\end{Verbatim}

\textbf{ConditionCode:}
\begin{Verbatim}
# (Already implemented above)
\end{Verbatim}

\textbf{ExampleFail:}
\begin{Verbatim}
Positive example not inferred (NONE description)
\end{Verbatim}

\textbf{Analysis:}
\begin{Verbatim}
Code correct, Example incorrect. No example was inferred. The code correctly implements constraints to disallow the same breeds on any two consecutive days.
\end{Verbatim}

\hrulefill

\textbf{InitialContext:}
\begin{Verbatim}
An administrator must assign parking spaces to six new employees: Robertson, Souza, Togowa, Vaughn, Xu, and Young. Each of the six employees must be assigned one of the following parking spaces: #1, #2, #3, #4, #5, or #6. No two employees can be assigned the same parking space.
\end{Verbatim}

\textbf{InitialCode:}
\begin{Verbatim}
from z3 import *

employees_sort, (Robertson, Souza, Togowa, Vaughn, Xu, Young) = EnumSort('employees', ['Robertson', 'Souza', 'Togowa', 'Vaughn', 'Xu', 'Young'])
parking_spaces_sort = IntSort()
parking_space = Function('parking_space', employees_sort, parking_spaces_sort)

pre_conditions = []

pre_conditions.append(Distinct([parking_space(e) for e in [Robertson, Souza, Togowa, Vaughn, Xu, Young]]))
pre_conditions.append(And([parking_space(e) >= 1 for e in [Robertson, Souza, Togowa, Vaughn, Xu, Young]]))
pre_conditions.append(And([parking_space(e) <= 6 for e in [Robertson, Souza, Togowa, Vaughn, Xu, Young]]))

\end{Verbatim}

\textbf{Condition:}
\begin{Verbatim}
If Young is assigned a higher-numbered parking space than Souza, then Young is assigned parking space #2.
\end{Verbatim}

\textbf{ConditionCode:}
\begin{Verbatim}
check_property = parking_space(Young) == 2
pre_conditions.append(check_property)
\end{Verbatim}

\textbf{ExampleFail:}
\begin{Verbatim}
Positive example not inferred (NONE description)
\end{Verbatim}

\textbf{Analysis:}
\begin{Verbatim}
Code incorrect, Example incorrect. The code does not implement the conditional requirement but simply asserts that Young should be assigned #2. No positive example is inferred.
\end{Verbatim}

\hrulefill

\textbf{InitialContext:}
\begin{Verbatim}
A bakery makes exactly three kinds of cookie—oatmeal, peanut butter, and sugar. Exactly three batches of each kind of cookie are made each week (Monday through Friday) and each batch is made, from start to finish, on a single day.
\end{Verbatim}

\textbf{InitialCode:}
\begin{Verbatim}
from z3 import *

cookies_sort, (oatmeal, peanut_butter, sugar) = EnumSort('cookies', ['oatmeal', 'peanut_butter', 'sugar'])
days_sort, (Monday, Tuesday, Wednesday, Thursday, Friday) = EnumSort('days', ['Monday', 'Tuesday', 'Wednesday', 'Thursday', 'Friday'])
batches_sort = IntSort()
made_on = Function('made_on', cookies_sort, batches_sort, days_sort)

pre_conditions = []

# Ensure that each kind of cookie has exactly three batches made on different days
for cookie in [oatmeal, peanut_butter, sugar]:
    pre_conditions.append(Distinct([made_on(cookie, b) for b in range(1, 4)]))

# Ensure that each batch number is between 1 and 3
for cookie in [oatmeal, peanut_butter, sugar]:
    for b in range(1, 4):
        pre_conditions.append(And(b >= 1, b <= 3))
\end{Verbatim}

\textbf{NegExample:}
\begin{Verbatim}
And(made_on(oatmeal, 1) == Monday, made_on(oatmeal, 2) == Tuesday, made_on(oatmeal, 3) == Wednesday)
\end{Verbatim}

\textbf{NegExampleDescription:}
\begin{Verbatim}
The oatmeal cookies are made on Monday, Tuesday, and Wednesday, which violates the constraint that each kind of cookie must be made on different days.
\end{Verbatim}

\textbf{Analysis:}
\begin{Verbatim}
Code incorrect, Example incorrect. The negative example does not violate the intended constraints as it simply assigns oatmeal batches to 3 different days. The code does not prevent any batch numbers higher than 3.
\end{Verbatim}

\subsection{Runtime performance and optimizations}
\label{app:runtimeanalysis}

We conducted an evaluation of the runtime performance of the current system. Executing the system over a sample of 250 data points (50 from each dataset), the median runtime per task is 152 seconds (around 2.5 minutes), with first quartile 108s, third quartile 267s and mean 249s. This was on an Intel Xeon Gold 6126 CPU @ 2.60 GHz with 16 cores and no hyper-threading, 62 GB of RAM, and an HDD-based storage system (this machine has slightly lower single-threaded performance than most modern desktops). However, there are also many potential optimizations to the SSV algorithm that can be made to significantly reduce the run time in a practical implementation:

\begin{itemize}
    \item The outer temperature loop (line 2 in Figure \ref{fig:ssv-alg}) can be fully parallelized as all the computations are independent for each temperature. That can yield up to 4X speed up (with 4 temperatures being tried in our current system). Side note: even with a single temperature of 0, our algorithm still beats all baselines in terms of accuracy (as in our ablation study), so even such an ablated system would be beneficial if computation costs are of significant concern.

    \item In the verification phase (line 9 in Figure \ref{fig:ssv-alg}), the solver calls to verify each of the concrete instantiations can be parallelized as they are checked independently. These are around 10 to 20 independent solver calls on average (2 instantiations each for around 5-10 constraints) that can be parallelized for significant speedup.

    \item Caching solver verification checks between repair attempts. Currently for each repair attempt in the inner loop (line 4 in Figure \ref{fig:ssv-alg}), we perform the full verification on the repaired program (on all constraints). However, most of the time the repaired change is on a single constraint for which a failing instantiation was found and all other constraints remain identical (though not always guaranteed as in some rare cases the LLM may reformulate the whole program). Hence if we cache the solver requests for each instantiation verification, many of these repetitive checks can be avoided in the repaired programs for the constraints that are unaltered.
\end{itemize}

As a general side note, recent reasoning-oriented models such as Open AI’s o1 can take several seconds or up to a few minutes on some tasks with significantly more computational resources/GPUs, so higher runtimes in the order of a few minutes may generally be expected to robustly address complex reasoning problems.

\end{document}